\documentclass[journal]{IEEEtran}
\usepackage{graphicx}
\usepackage{booktabs}
\usepackage{multirow}
\usepackage{amsmath}
\usepackage{amssymb}
\usepackage[caption=false, font=footnotesize]{subfig}
\usepackage[dvipsnames]{xcolor}
\usepackage{threeparttable}
\usepackage{hyperref}
\usepackage[normalem]{ulem}
\usepackage{xcolor}
\newcommand{\revise}[1]{}

\ifCLASSINFOpdf
\else
\fi

\begin{document}

\title{HTC-DC Net: Monocular Height Estimation from Single Remote Sensing Images}

\author{Sining~Chen, 
        Yilei~Shi,~\IEEEmembership{Member,~IEEE,}
        Zhitong~Xiong,~\IEEEmembership{Member,~IEEE,}\\
        and~Xiao~Xiang~Zhu,~\IEEEmembership{Fellow,~IEEE}\\
\thanks{The work is jointly supported by the German Federal Ministry of Education and Research (BMBF) in the framework of the international future AI lab "AI4EO -- Artificial Intelligence for Earth Observation: Reasoning, Uncertainties, Ethics and Beyond" (grant number: 01DD20001) and by German Federal Ministry for Economic Affairs and Climate Action in the framework of the "national center of excellence ML4Earth" (grant number: 50EE2201C), by the Munich Center for Machine Learning and by the TUM Georg Nemetschek Institute for Artificial Intelligence for the Built World as part of the AI4TWINNING project. The contributions of S. Chen was carried out in part during the time when he was jointly affiliated with the Technical University of Munich and the German Aerospace Center supported by a DAAD scholarship.}
\thanks{Corresponding author: Xiao Xiang Zhu.}
\thanks{S. Chen, Z. Xiong, and X. Zhu are with the Chair of Data Science in Earth Observation, Technical University of Munich (TUM), 80333 Munich, Germany (e-mail: sining.chen@tum.de; zhitong.xiong.tum.de; xiaoxiang.zhu@tum.de). 
}
\thanks{X. Zhu is also with the Munich Center for Machine Learning.}
\thanks{Y. Shi is with the School of Engineering and Design, Technical University of Munich (TUM), 80333 Munich, Germany (e-mail: yilei.shi@tum.de).}
\thanks{}
}
\markboth{submitted to IEEE Transactions on Geoscience and Remote Sensing}%
{Shell \MakeLowercase{\textit{et al.}}: Bare Demo of IEEEtran.cls for Journals}

\maketitle

\begin{abstract}
3D geo-information is of great significance for understanding the living environment; however, 3D perception from remote sensing data, especially on a large scale, is restricted, mainly due to the high costs of 3D sensors such as LiDAR. To tackle this problem, we propose a method for monocular height estimation from optical imagery, which is currently one of the richest sources of remote sensing data. As an ill-posed problem, monocular height estimation requires well-designed networks for enhanced representations to improve performance. Moreover, the distribution of height values is long-tailed with the low-height pixels, e.g., the background, as the head, and thus trained networks are usually biased and tend to underestimate building heights. To solve the problems, instead of formalizing the problem as a regression task, we propose HTC-DC Net following the classification-regression paradigm, with the head-tail cut (HTC) and the distribution-based constraints (DCs) as the main contributions. HTC-DC Net is composed of the backbone network as the feature extractor, the HTC-AdaBins module, and the hybrid regression process. The HTC-AdaBins module serves as the classification phase to determine bins adaptive to each input image. It is equipped with a vision transformer encoder to incorporate local context with holistic information and involves an HTC to address the long-tailed problem in monocular height estimation for balancing the performances of foreground and background pixels. The hybrid regression process does the regression via the smoothing of bins from the classification phase, which is trained via DCs. The proposed network is tested on \revise{three} datasets of different resolutions, namely \revise{ISPRS Vaihingen (0.09 m)}, DFC19 (1.3 m) and GBH (3 m). Experimental results show the superiority of the proposed network over existing methods by large margins. Extensive ablation studies demonstrate the effectiveness of each design component. Codes and trained models are published at \url{https://github.com/zhu-xlab/HTC-DC-Net}.
\end{abstract}


\IEEEpeerreviewmaketitle
\section{Introduction}
\IEEEPARstart{M}{onocular} height estimation is the process of deriving height information from single remote sensing images. The generated height maps, usually delivered in the form of digital surface models (DSMs) or normalized digital surface models (nDSMs), are of great importance for many downstream applications. For example, estimating building heights is essential for 3D building models \cite{arefi2013, partovi2014, wang2021}, which serve as a crucial information basis for urban planning and disaster management. And modeling vegetation heights, represented as canopy height models \cite{matese2017, ota2015, sadeghi2016}, can improve the understanding of biomass and, thus, help the carbon cycle studies on a large scale.

Height information can be retrieved directly with 3D-aware techniques, e.g., 3D sensors such as light detection and ranging (LiDAR) \cite{priestnall2000, elaksher, chen2022} and synthetic aperture radar (SAR) \cite{zhu2014, zhu2014a}, or stereo pairs of optical images. However, such techniques are only conditionally applicable for various reasons. Though LiDAR delivers high-quality 3D measurements, the very high operational costs hamper its use in most cases. SAR has larger coverage than LiDAR, and satisfactory accuracy, however, suffers a lot from the side-looking geometry \cite{sun2022b, sun2022c}. In dense urban areas, the need for a stack of SAR images restricts its applicability for 3D reconstruction in practice \cite{shi2020}. While stereo images are easier to obtain, the compromise between acquisition quality and quantity poses great difficulties \cite{tack2012,ginzler2015}. Large-scale applications demand high-quality and comprehensive data, which is not adequately met by either costly aerial imaging acquisitions or low-budget satellite stereo pairs. While the former provides high-quality data, it comes at a significant expense, whereas the latter is typically affected by cloud contamination and long baselines, limiting its usefulness for large-scale applications \cite{xiong2021benchmark}.

In contrast, monocular images, especially those from satellites, are rich in quantity \cite{xiong2022}, which addresses the deficiencies of the aforementioned techniques and, thus, can support large-scale applications as well as the corresponding updates. The only problem is how to mine the concealed height information from them. Early works on monocular height estimation focus on the level of instances \cite{huang2007,izadi2012}. Following the physical model of shadow casting, shadow lengths are exploited as the cue for inferring heights. Together with solar parameters, the heights of ground objects can be computed mathematically. However, such methods suffer from overlaps between shadows and objects, especially in dense urban areas or dense forests, as well as the availability of exact solar parameters. Fortunately, a large amount of data and the recent emergence of deep learning methods make it possible to tackle the problem in a data-driven manner. Given that sufficient data could be used for training, models of high performance could be expected.

Monocular height estimation could be inspired by advances in monocular depth estimation \cite{zhao2020}, which is faced with exactly the same problem as monocular height estimation, that being the ill-posed nature. Namely, multiple height maps with similar height structures could look very similar in the domain of optical images; thus, one specific optical image can correspond to multiple height map predictions that are hard to disentangle. Inspired by the use of vision transformers (ViTs) \cite{dosovitskiy2021} for enforcement of global consistency in monocular depth estimation, we propose involving a ViT for modeling long-range attention to combat the ill-posed problem. Besides, changes in solution paradigms have occurred in monocular depth estimation. Instead of solving the problem as a regression task, the state-of-the-art solution is to convert the problem into a classification-regression problem. For example, the hybrid regression process \cite{bhat2021} is proposed to facilitate the solution for monocular depth estimation. In this paper, we demonstrate the feasibility and superiority of applying the classification-regression paradigm for monocular height estimation from remote sensing images.

Different from monocular depth estimation, monocular height estimation also suffers from the long-tailed distribution problem \cite{zhang2021}. Specifically, in the physical world, most of the ground objects are of lower height, such as low buildings and vegetation, while high objects are rare, e.g., skyscrapers. When a network is trained with such data from nature, it will be largely biased. Considering that the long-tailed distribution of height values is even more skewed than the worst cases in long-tailed classification, the predictions can include many fatal cases for higher objects, with incredibly large errors. 

Different from a simple regression process, the hybrid regression process incorporates a distribution-based approach, which yields a distribution specified by the bin centers from the classification phase and the bin probabilities from the regression phase. Theoretically, the final prediction lies within the bins with the highest probabilities. In practice, to avoid discrete predicted values, the final prediction is computed as the weighted average of the bin centers according to the bin probabilities, equivalent to the expectation value of a distribution. This is based on the assumption that the expectation value of the distribution is close to the value where the probability is the highest. However, the assumptions cannot be guaranteed without any constraints set on the distribution.

To cope with the aforementioned problems, we propose HTC-DC Net, which is equipped with a head-tail cut (HTC) and distribution-based constraints (DCs).

In summary, our contributions are as follows:
\begin{itemize}
    \item We propose a novel architecture for monocular height estimation. We utilize a classification-regression paradigm for HTC-DC Net, which employs the ViT for enhanced representation learning.
    \item We propose an HTC to address the extremely long-tailed nature of height values, i.e., to mitigate the side impact of the background pixels as the majority.
    \item We propose using DCs to regularize the bin probabilities used during the regression phase, which are mathematically neat and lead to remarkable improvements.
    \item We conduct extensive experiments to showcase the efficacy of the proposed network and comprehensive ablation studies to demonstrate the necessity of each designed component. The proposed network outperforms the existing methods by a large margin.
\end{itemize}

The remainder of the article is organized as follows. Section \ref{chap:related_works} gives an overview of the related works. The proposed method is described in detail in Section \ref{chap:methodology}, followed by Section \ref{chap:experiments} describing the experiments and Section \ref{chap:results} showing the experimental results. Discussions and ablation studies are presented in Section \ref{chap:discussions}. Conclusions are drawn, and further research directions are described in Section \ref{chap:conclusions}.
\section{Related Works}\label{chap:related_works}
\begin{figure*}[t]
    \centering
    \includegraphics[width=\linewidth]{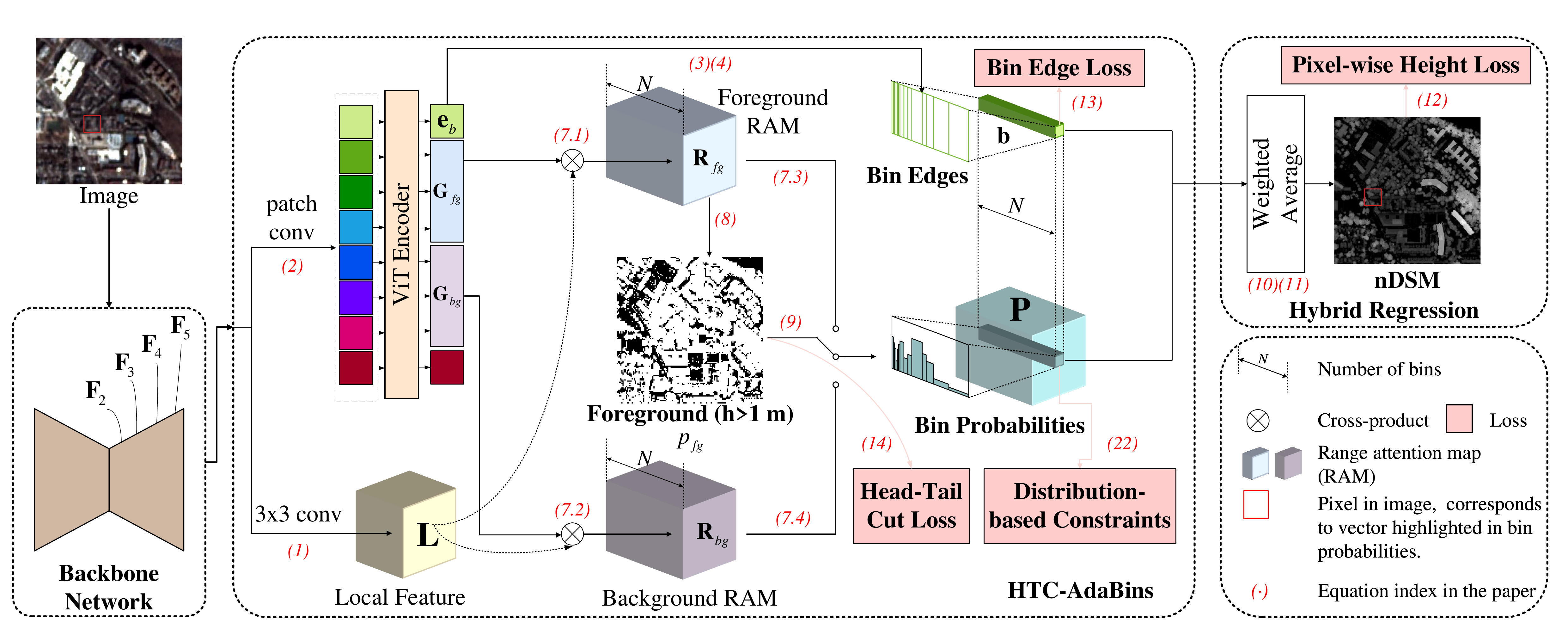}
    \caption{Network Architecture of HTC-DC Net. Following the classification-regression paradigm, the HTC-DC Net is formalized into three parts, the backbone network, the HTC-AdaBins module as the classification phase, and the hybrid regression process as the regression phase. First, a backbone network is used to extract features from images. Based on the features, the HTC-AdaBins module derives the bin edges, which serve as the discretization of the height value range into adaptive bins as classes, and the bin probabilities, regarded as the class probabilities. Finally, the hybrid regression process converts the discretized output space back to a continuous output space by a weighted average of the bin centers according to the bin probabilities. As the contributions of the paper, the head-tail cut (HTC) in the HTC-AdaBins module is used to treat foreground and background pixels separately to account for the long-tail effect in monocular height estimation, and distribution-based constraints (DC) are applied to the predicted bin probabilities for regularization. The foreground refers to pixels higher than 1 m. The red numbers in parentheses refer to the corresponding equations.} 
    \label{fig:architecture}
\end{figure*}
\subsection{Monocular Height Estimation}
Deep-learning-based monocular height estimation networks can be categorized into pixel-wise and instance-wise methods, based on their distinct objectives. As the major focus of the paper, pixel-wise height estimation can be formalized as a deep dense regression task. To tackle the task, encoder-decoder fully convolutional networks (FCNs) are mostly utilized \cite{shelhamer2017}. FCNs for semantic segmentation can be adopted by removing the final classification layer (Softmax or Sigmoid activation), e.g., SegNet \cite{badrinarayanan2017}, U-Net \cite{ronneberger2015}, Eff-UNet \cite{baheti2020}. Besides, many networks are proposed specifically for monocular height estimation. Those networks can be categorized into two types: single-task learning networks and multi-task learning networks. 
\subsubsection{Single-task Learning Networks}
Single-task learning networks map the input images to the output height maps. In such networks, feature fusion is usually applied to boost performance. For instance, Mou \textit{et al.} \cite{mou2018} proposed one of the first deep-learning-based methods to estimate height from a single optical image---an encoder-decoder neural network, IM2HEIGHT. Compared to plain FCN, IM2HEIGHT has a skip connection, accounting for low-level feature fusion. It leads to sharper object edges and more details preserved. Amirkolaee \textit{et al.} \cite{amirkolaee2019} adopted advanced techniques from the CV community. They used the up-sampling block to lighten the computation burden and utilized multi-level feature fusion to combat the blurring effect during inference \cite{laina2016}. Besides, they also proposed a post-processing scheme to enforce the continuities around the patch edges. Xing \textit{et al.} \cite{xing2021} proposed PLNet with the feature fusion module---gated feature aggregation module (GFAM) and a refining module---progressive refine module (PRM).
\subsubsection{Multi-task Learning Networks}
Multi-task learning networks introduce auxiliary tasks in addition to height predictions, with the expectation that both tasks support each other during training. Usually, based on the assumption that heights and semantics are highly correlated \cite{xiong2022a}, semantic segmentation can be regarded as an auxiliary task for height estimation. For example, Srivastava \textit{et al.} \cite{srivastava2017} were the first to showcase the gains the auxiliary semantic segmentation head brings. Carvalho \textit{et al.} \cite{carvalho2020} explored earlier separation between heights and semantics, and compared different multi-task learning strategies. Elhousni \textit{et al.} \cite{elhousni2021} used further auxiliary geometric information, the normal vectors, in a two-stage network, where the first stage results are fed into the second stage de-noising autoencoder for refinement.

As an alternative, monocular height estimation can also be regarded as an image translation task, assuming that the images and heights are backed by the same underlying semantics. In this context, generative adversarial networks (GANs) are used. Ghamisi and Yokoya \cite{ghamisi2018} used a GAN-based network consisting of a generator and a discriminator. The generator applies the style transfer, i.e., takes the input image, and transfers it to the output height map. The discriminator is used to help train the generator to generate realistic height maps. The network is trained with image-height map pairs. Later, Paoletti \textit{et al.} \cite{paoletti2021} overcame the problem by introducing the shared latent features. This makes the network generalize better and learn more generic style information. Improvements in the performances are demonstrated by experiments.

Besides, methods for instance-wise monocular height estimation, though not the focus of the paper, account for cases when doing 3D perception specifically for some ground objects, e.g., buildings. Under such circumstances, instance segmentation-based networks can be used, where heights are predicted conditioned on the instances as the prior. By this means, the output maps are usually sparse maps with only object pixels filled with object-wise single height values, which, in the context of 3D building reconstruction, are exactly the LoD-1 (Level of Details) building models. Such methods are usually based on two-stage instance segmentation networks, e.g., Mask R-CNN \cite{he2018}. Mahmud \textit{et al.} \cite{mahmud2020} modified Mask R-CNN into a multi-task network, with a joint prediction of heights, signed distance function, and semantics aggregated into the final output. Chen \textit{et al.} \cite{chen2021} proposed a network named as Mask-Height R-CNN, which is adapted from Mask R-CNN for monocular height estimation by adding a height regression head to the region proposal network (RPN) \cite{sun2022a}. Recently, Li \textit{et al.} \cite{li2023} proposed a novel type of representation for building instances in 3D space: 3D centripetal shift representation. Their proposed network, termed as 3DCentripetalNet, learns 3D centripetal shift representation and building corners, which are further utilized to retrieve building heights.
\subsection{Monocular Depth Estimation}
As a highly related task to monocular height estimation, monocular depth estimation has been a long-standing task in the computer vision community \cite{zhao2020}. The advances in monocular depth estimation can thus inspire better solutions for monocular height estimation. It has been witnessed that the paradigm of doing monocular depth estimation changes from regression to classification, then to classification-regression---the state-of-the-art solution.

Treated as a regression problem, the depths are predicted directly from the images, which are supervised by the ground truth depth values. There have been many works in this direction, and they are all intuitive; however, their performances are limited \cite{eigen2014,eigen2015,laina2016,zhang2018,lore2018}. Fu \textit{et al.} \cite{fu2018} proposed using ordinal regression, which converts the regression problem into a classification problem, which inspired the application of DORN in MHE \cite{li2020}. Besides, Sun \textit{et al.} \cite{sun2022a} designed a classification network based on the ordinal regression network, however, with an adaptive bin design and a set prediction framework for bin prediction. The results of these classification methods are discrete with artifacts; however, the overall performances are better than the regression networks. Recently, the classification-regression scheme has emerged with state-of-the-art performances \cite{bhat2021,li2022,bhat2022}. They propose to use adaptive bins learned from the image to reflect the real distribution of ground truth values of each image and then predict the depth values by a weighted average of the learned bins. Compared to ordinal regression networks, such as DORN \cite{li2020}, classification-regression networks can adapt to different input images and output continuous depth maps.
\section{Methodology}\label{chap:methodology}
\begin{figure*}[!t]
	\centering
	\includegraphics[width=\linewidth]{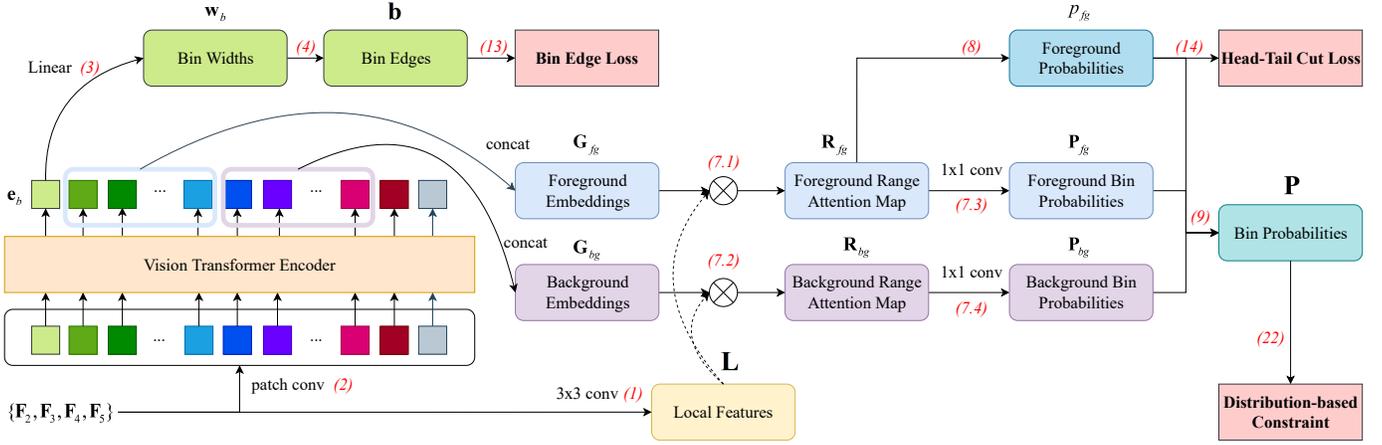}
	\caption{The HTC-AdaBins module contains two branches, namely the local branch and the global branch. The local branch is responsible for computing local features with a convolutional layer, while the global branch is involved with a vision transformer encoder for capturing global context. The embeddings from the ViT encoder are utilized for computing the bin edges, the foreground bin probabilities, and the background bin probabilities, respectively. During the computation of bin probabilities, a cross-product of the local features and the embeddings from the ViT encoder is conducted to incorporate features of different scopes. The head-tail cut is derived from the foreground range attention map and used to combine bin probabilities maps for foreground and background pixels. The outputs of the HTC-AdaBins module are supervised by a bin edge loss, a head-tail cut loss, and a distribution-based constraint. The red numbers in parentheses refer to the corresponding equations.}
	\label{fig:adabins}
\end{figure*}
As shown in Fig. \ref{fig:architecture}, the proposed HTC-DC Net consists of three parts, the backbone network to extract features from input images, the HTC-AdaBins module to conduct the HTC as well as incorporate local and holistic information, and the hybrid regression module to get the final height predictions. The proposed network follows the classification-regression paradigm: Based on the extracted features, the HTC-AdaBins module conducts the classification of pixels into bins that are adaptive to each input image, and the hybrid regression module smooths the discrete bins into the continuous output space. The three components are described in detail in this section.
\subsection{Backbone Network}
Instead of directly using an encoder-decoder structure for height prediction, the backbone network is used for the generation of feature maps $\{\mathbf{F}_1, \mathbf{F}_2, \mathbf{F}_3, \mathbf{F}_4, \mathbf{F}_5\}$ from input images $\mathbf{I}\in\mathbb{R}^{3\times H_0\times W_0}$, which contain rich spatial and spectral information. Inspired by networks, e.g., U-Net \cite{ronneberger2015}, where the intermediate features are aggregated in the later stages of the networks, and following the advanced design in hybrid regression for monocular depth estimation \cite{li2022,bhat2022}, the early injection is done by applying the HTC-AdaBins module and the following hybrid regression process to features of multiple stages in the decoder network, resulting in predictions at different scales. The results of intermediate levels are not taken as the final output, however, used for the computation of loss functions for training.
\subsection{HTC-AdaBins}\label{sec:adabins}
The HTC-AdaBins module (see Fig. \ref{fig:adabins}) is a variant of the AdaBins module \cite{bhat2021} with modifications to address the long-tailed distribution problem in monocular height estimation from remote sensing images. It is used to obtain bin edges $\mathbf{b}\in\mathbb{R}^{N+1}$ and bin probabilities $\mathbf{P}\in\mathbb{R}^{N\times H\times W}$ from the feature maps generated by the backbone network $\mathbf{F}\in\mathbb{R}^{C\times H\times W}$, where $N$ is the number of bins as a hyperparameter, the same for all input images. Intuitively, the bins discretize the continuous height into classes, which are adaptive to each input image by reflecting the height value distribution of each image, and the bin probabilities serve as the class probabilities. That is, the HTC-AdaBins module converts the regression problem into a classification problem. Besides, the HTC-AdaBins module enables the interaction between local textures learned by the local branch and the global context learned by the global branch. In addition, the HTC enables different treatment of the foreground and the background pixels, such that the performances for foreground and background pixels are balanced.
\subsubsection{Local and Global Branch}
The local branch with one convolutional layer exploits the local feature pattern $\mathbf{L}$ as
\begin{equation}
    \mathbf{L}=\text{conv}_{3\times 3}(\mathbf{F}).
\end{equation}
While the global branch with a ViT encoder \cite{dosovitskiy2021} models the global context. To be fed into the global branch, the feature maps are divided into patches, among which the relations are modeled to refine the embeddings. The process is denoted as
\begin{equation}
    \mathbf{E}=\displaystyle\{\mathbf{e}_1, \mathbf{e}_2, \cdots, \mathbf{e}_{\frac{HW}{p^2}}\}=\text{ViT}(\text{conv}_{p\times p}(\mathbf{F})),
\end{equation}
where $p$ denotes the patch size. The resulting embeddings $\mathbf{E}$ are taken for different uses. To obtain the bins, the first embedding $\mathbf{e}_b:=\mathbf{e}_1$ is regarded as the bin width embedding. It is fed into a linear layer fc, then normalized by a softmax function to get the relative bin widths $\mathbf{w}_b\in\mathbb{R}^N$, defined as
\begin{equation}\label{eqn:bin_width}
    \mathbf{w}_b=\text{softmax}(\text{fc}(\mathbf{e}_b)).
\end{equation}
Finally, given the minimal and the maximal possible values of heights, $h_{\text{min}}$ and $h_{\text{max}}$, the bin edges $\mathbf{b}$ can be obtained by
\begin{equation}\label{eqn:bin_edges}
    \begin{array}{r@{}l}
        \mathbf{b}_0&=h_{\text{min}}, \\
        \mathbf{b}_i&=\mathbf{b}_{i-1}+\mathbf{w}_{bi}, \forall i=1,2,\cdots,N.
    \end{array}
\end{equation}

A fixed number $m$ of the embeddings following the first one are concatenated and taken as the global feature $\mathbf{G}:=\text{concatenate}\,(\mathbf{e}_2, \mathbf{e}_3, \cdots, \mathbf{e}_{m+1})$.
The global feature $\mathbf{G}$ from the global branch is incorporated with the output from the local branch by a cross-product as follows,
\begin{equation}\label{eqn:ram}
    \mathbf{R}=\mathbf{L}\times\mathbf{G},
\end{equation}
to compute the range attention maps (RAMs) $\mathbf{R}$, which represent the extent of how the height value distribution of a local area compares to the global distribution. 

The RAMs $\mathbf{R}$ are then convolved and normalized to get the bin probability maps $\mathbf{P}$ as
\begin{equation}\label{eqn:bp}
    \mathbf{P}=\text{softmax}(\text{conv}_{1\times 1}(\mathbf{R})).
\end{equation}
\subsubsection{Head-Tail Cut to Combat the Long-Tailed Effect}
\begin{figure}[t]
    \centering
    \includegraphics[width=\linewidth]{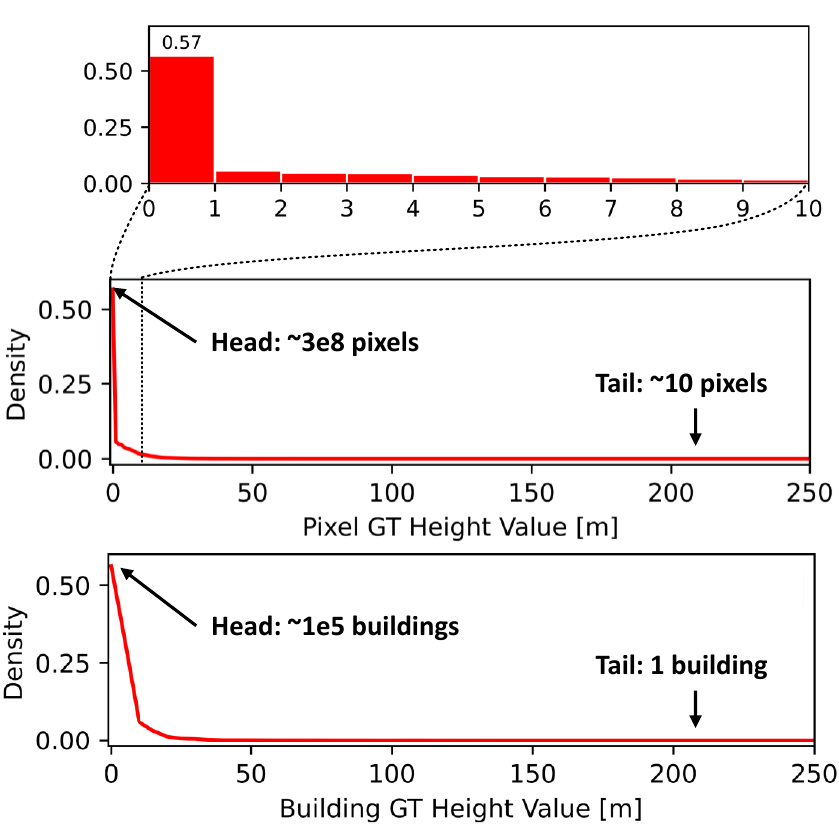}
    \caption{Height value distribution of GBH training and validation set. The background with height values smaller than 1 m consists of around 3e8 pixels, which accounts for 57\% of the total pixels, while the pixels with very large height values only count to approximately 10 for each 1 m bin. The long-tailed distribution also exists in building height values.}
    \label{fig:dis}
\end{figure}
As mentioned above, in remote sensing, the height values are usually extremely long-tailed distributed (see Fig. \ref{fig:dis}), so the majority background pixels may disturb the computation of RAMs. To mitigate this effect, we propose using an HTC to separate the computation of foreground and background pixels, where foreground pixels are defined as pixels with height values greater than 1 m. The definition of the threshold is proved to be reasonable through experiments.

The separation takes effect within the ViT encoder from the global branch. Instead of computing a unique RAM, two different sets of tokens of the same number $m$, $\mathbf{G}_{fg}=\text{concatenate}\,(\mathbf{e}_2, \mathbf{e}_3, \cdots, \mathbf{e}_{m+1})$, and $\mathbf{G}_{bg}=\text{concatenate}\,(\mathbf{e}_{m+2}, \mathbf{e}_{m+2}, \cdots, \mathbf{e}_{2m+1})$, are selected to compute the foreground and background RAMs, $\mathbf{R}_{fg}\in \mathbb{R}^{N\times H\times W}$ and $\mathbf{R}_{bg}\in \mathbb{R}^{N\times H\times W}$, and then the bin probabilities $\mathbf{P}_{fg}$ and $\mathbf{P}_{bg}$, respectively. The Eqn. \ref{eqn:ram} and Eqn. \ref{eqn:bp} are rewritten as
\begin{equation}\label{fig:htc_rp}
    \begin{array}{r@{}l}
        \mathbf{R}_{fg}&=\mathbf{L}\times\mathbf{G}_{fg},\\
        \mathbf{R}_{bg}&=\mathbf{L}\times\mathbf{G}_{bg},\\
        \mathbf{P}_{fg}&=\text{softmax}(\text{conv}_{1\times 1}(\mathbf{R}_{fg})),\\
        \mathbf{P}_{bg}&=\text{softmax}(\text{conv}_{1\times 1}(\mathbf{R}_{bg})).
    \end{array}
\end{equation}
In this way, on the one hand, the embeddings from the ViT are utilized in a more efficient way so as to take full advantage of the holistic information acquired by the great computational effort, and on the other hand, foreground and background are perceived earlier in the global attention phase, resulting in more distinguishable treatments for them.

The extreme height distribution in the physical world renders the foreground and background pixels each about half of the whole dataset. The HTC problem makes an almost balanced binary classification setting, which is done by simply adding a binary classification head on the foreground RAMs. The probability that pixels belong to the foreground is computed as
\begin{equation}
    p_{fg}=\text{sigmoid}(\mathbf{R}_{fg}).
\end{equation}
The probability map $p_{fg}$ serves as a mask to combine the bin probabilities computed for foreground and background pixels, written as
\begin{equation}
    \mathbf{P}=(p_{fg}>0.5)\cdot\mathbf{P}_{fg}+(p_{fg}\leq 0.5)\cdot\mathbf{P}_{bg}.
\end{equation}
\subsection{Hybrid Regression Process}
The hybrid regression process is designed to combine the learned information for each bin by smoothing the discrete output space derived from the HTC-AdaBins module to a continuous output space.

First, a representative value from each bin, i.e., the bin center $\mathbf{c}$, is calculated as the midpoints between two bin edges with
\begin{equation}\label{eqn:bin_center}
    \mathbf{c}_i=\frac{\mathbf{b}_{i-1}+\mathbf{b}{i}}{2}, \forall i=1,2,\cdots,N.
\end{equation}
The final predicted height map $\mathbf{H}\in\mathbb{R}^{1\times H\times W}$ is formalized as a weighted average of the $N$ bin centers $\mathbf{c}$ according to the bin probabilities $\mathbf{P}$, i.e.,
\begin{equation}\label{eqn:hr}
    \mathbf{H}=\sum_i^N \mathbf{P}_i\mathbf{c}_i.
\end{equation}
\subsection{Loss Functions}
The loss function is composed of four parts.
\subsubsection{Pixel-wise Height Loss}
The pixel-wise height loss is defined as the L1 loss, written as
\begin{equation}\label{eqn:height_loss}
    \mathcal{L}_h=\frac{1}{|\mathbf{H}|}\sum L_1(\mathbf{H}, \tilde{\mathbf{H}}),
\end{equation}
where $|\mathbf{H}|$ denotes the total number of pixels, and $\tilde{\mathbf{H}}$ denotes the ground truth height map.
\subsubsection{Bin Edge Loss}
To make sure the bin edges comply with the distribution of ground truth values, Chamfer loss \cite{fan2016}, which computes the bi-directional distances between two point sets, is utilized to supervise the bin edge predictor, i.e.,
\begin{equation}\label{eqn:bin_loss}
    \mathcal{L}_b=\text{chamfer}(\mathbf{b}, \text{flatten}(\tilde{\mathbf{H}})).
\end{equation}
Intuitively, the bin edges and the flattened ground truth height maps are seen as two 1D point sets with height values as the coordinates. For each point in one set, the closest point in the other set is searched, and the distance between the two points is computed and added to the final loss. On the one hand, the distances to the bin edges force them to come from the height values of the input images. On the other hand, the distances to the height values of pixels encourage the bin edges to spread according to the distribution of pixel height values. When the Chamfer loss is small, the distances between the two point sets are small, i.e., the locations of the bin edges comply with the distribution of height values. Ideally, the bin edges lie at the quantiles of the ground truth height values.
\subsubsection{Head-Tail Cut Loss}
As mentioned in Section \ref{sec:adabins}, an HTC is conducted in the AdaBins module by a binary classification head. The HTC is supervised by a cross-entropy loss, denoted by
\begin{equation}\label{eqn:ce_loss}
	\mathcal{L}_{htc}=\text{cross-entropy}(p_{fg}, ~\tilde{\mathbf{H}}>1).
\end{equation}
\subsubsection{Distribution-based Constraint}\label{sec:dbc}
\begin{figure}[t]
	\centering
	\includegraphics[width=\linewidth]{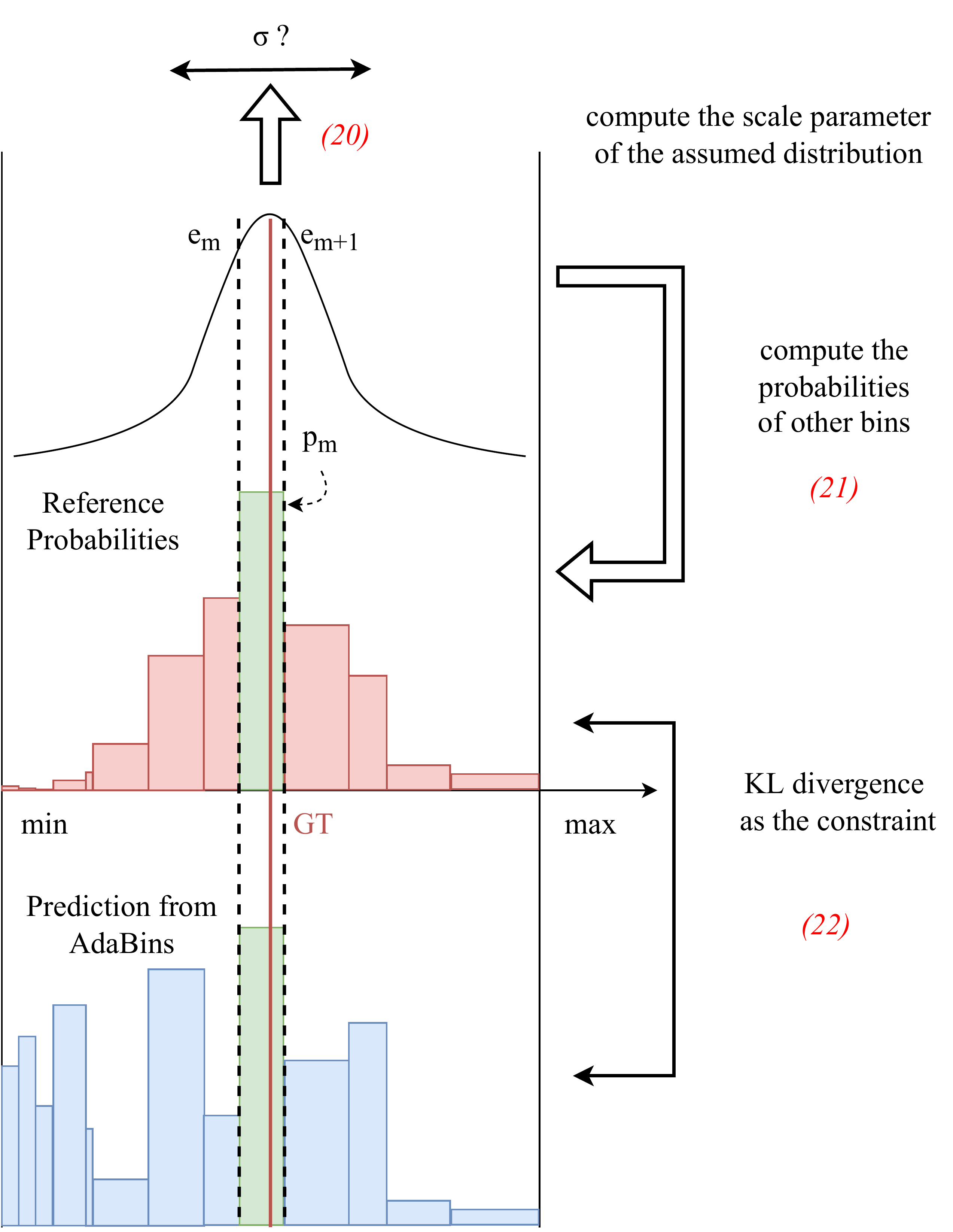}
	\caption{Distribution-based constraint. The constraint is derived and applied in three steps. First, the assumed distribution is solved from the probability of the bin where the GT value lies (the bin bounded by $e_m$ and $e_m+1$), which is assumed as the mode probability ($p_m$, in {\color{Green}green}). With the GT value taken as the mean value, the scale parameter, e.g., the standard deviation of a Gaussian distribution ($\sigma$), can be solved analytically. Second, following the assumed distribution, the probabilities for other bins can be calculated as the integral of the PDF within each bin and regarded as the reference bin probabilities (in {\color{Red}red}). Last, the reference bin probabilities (in {\color{Red}red}) are applied as a constraint for the predicted bin probabilities (in {\color{Blue}blue}) in the form of KL divergence. The red numbers in parentheses refer to the corresponding equations.}
	\label{fig:probloss}
\end{figure}
Conventionally, the single regression process yields a single point estimation as the height prediction. In contrast, the hybrid regression process incorporates a distribution-based approach. In the classification phase, a distribution is specified by the bin centers and the bin probabilities. Trivially, taking the most probable value drawn from the distribution accounts for the final height prediction. However, it would lead to discrete output maps solely with values from the bin centers. To overcome this limitation, in the regression phase, a weighted average (Eqn. \ref{eqn:hr}) of bin centers according to the bin probabilities serves as the smoothing of the bins and, thus, enables continuous output space and approaches the mode of the distribution. From a probabilistic perspective, the weighted average is equivalent to computing the expectation value of the underlying height value distribution, which could be far from the distribution mode without any constraints on the distribution. One special case when the expectation and the mode of a distribution are close to each other is when the distribution is symmetric and unimodal, such as a Gaussian distribution. In this case, the hybrid regression process yields the mode value, approximated by the expectation value of the underlying distribution. 

Therefore, to enforce that the resulting expectation value of the distribution approaches the distribution mode, a DC is posed on the bin probabilities. As illustrated in Fig. \ref{fig:probloss}, if the predicted height values obey certain distributions, then the bins are intervals within the defined domain of the distribution, and the bin probabilities are the integrals within the bin intervals. Assuming a known distribution, such as a Gaussian distribution with the ground truth value as the mode, the whole distribution can then be computed and serve as a constraint.

Mathematically, consider that the predicted height $h$ for a pixel is subject to a Gaussian distribution, centered at the ground truth height value $\tilde{h}$, i.e.,
\begin{equation}
	h\sim\mathcal{N}(h|\tilde{h},\sigma^2)=\displaystyle\frac{1}{\sigma\sqrt{2\pi}}\exp(-\frac{(h-\tilde{h})^2}{2\sigma^2}),
\end{equation}
where $\sigma$ as the standard deviation specifies the scale of the distribution, which is unknown and to be solved. The corresponding cumulative distribution function $F$ is
\begin{equation}
	F(h)=\frac{1}{2}(1+\text{erf}(\frac{h-\tilde{h}}{\sigma\sqrt{2}})),
\end{equation}
where erf stands for the Gaussian error function, written as
\begin{equation}
    \text{erf}(x)=\frac{2}{\sqrt{\pi}}\int_0^x\exp(-t^2)\text{d}t.
\end{equation}
Then the probability for the bin where the ground truth value lies, i.e., the mode probability $P_m$, can be represented as
\begin{equation}\label{eqn:mode_prob}
	\begin{array}{r@{}l}
		P_m&=F(e_{m+1})-F(e_m)\\
		&=\displaystyle\frac{1}{2}(\text{erf}(\frac{e_{m+1}-\tilde{h}}{\sigma\sqrt{2}})-\text{erf}(\frac{e_m-\tilde{h}}{\sigma\sqrt{2}})),
        \end{array}
\end{equation}
where the edges $e_m$ and $e_{m+1}$ make the bin around the ground truth value. When the mode probability is assumed as the reference, Eqn. \ref{eqn:mode_prob} has a unique solution of the standard deviation $\sigma$, and then the distribution is fixed. The above-mentioned equation can be solved numerically by optimization; however, it leads to large computational burdens. Alternatively, to ease the computation and to solve the equation analytically, it is assumed that the ground truth value lies exactly at the bin center. Then Eqn. \ref{eqn:mode_prob} is simplified as
\begin{equation}
	P_m=\text{erf}(\frac{e_{m+1}-e_m}{\sigma 2\sqrt{2}}).
\end{equation}
Then $\sigma$ can be solved as
\begin{equation}
	\sigma=\frac{e_{m+1}-e_m}{2\sqrt{2}\text{ierf}(P_m)},
\end{equation}
where ierf is the inverse function of the Gaussian error function erf. Even though the assumption is not true, it is necessary to make the problem tractable. Surprisingly, the simplification still brings performance improvements. 

After the distribution is fixed by solving the standard deviation $\sigma$, the bin probabilities for other bins can be easily computed by
\begin{equation}\label{eqn:bin_prob}
	P_i=F(e_{i+1})-F(e_i).
\end{equation}
Then, these computed probabilities from the assumed distribution are used to supervise the prediction of bin probabilities by the Kullback--Leibler (KL) divergence. The loss can be formulated as
\begin{equation}\label{eqn:hr_loss}
\mathcal{L}_{dist}=\frac{1}{|\mathbf{P}|}\sum -\tilde{\mathbf{P}} \log\frac{\tilde{\mathbf{P}}}{\mathbf{P}},
\end{equation}
where $\tilde{\mathbf{P}}$ is the probability map from the assumed underlying distribution, with the probability for each pixel in each bin computed by Eqn. \ref{eqn:bin_prob}.

To consider the differences between foreground and background pixels, different distributions are assumed for them. For background pixels, the predicted height values $h$ are assumed to follow a uniform distribution, with the ground truth value $\tilde{h}$ as the center point, written as
\begin{equation}
    h\sim\text{Uniform}(h|a,b)=\frac{1}{b-a},
\end{equation}
where $a$ and $b$ are the lower and upper bounds of the distribution. Similarly, to solve the distribution parameters, the mode probability $P_m$ is set as the reference:
\begin{equation}
    P_m=F(e_{i+1})-F(e_i)=\frac{e_{i+1}-e_{i}}{b-a}.
\end{equation}
The scale parameter of the distribution denoted as the width $w:=b-a$ is derived as
\begin{equation}
    w=\frac{e_{i+1}-e_i}{P_m}.
\end{equation}
Given the distribution is centered at the ground truth height value $\tilde{h}$, then the parameters can be derived as
\begin{equation}
    \begin{array}{r@{}l}
         a&=\tilde{h}-\displaystyle\frac{e_{i+1}-e_i}{2P_m},  \\
         b&=\tilde{h}+\displaystyle\frac{e_{i+1}-e_i}{2P_m}. 
    \end{array}
\end{equation}
Finally, the probabilities for other bins can be easily computed using Eqn. \ref{eqn:bin_prob}.

To compute the total loss, for a single feature map $\mathbf{F}_i$ of level $i$, the loss function $\mathcal{L}_i$ is defined as the summation of the aforementioned loss function parts, denoted by
\begin{equation}\label{eqn:total_loss}
    \mathcal{L}_{i}=\mathcal{L}_h+\mu_1\mathcal{L}_b+\mu_2\mathcal{L}_{htc}+\mu_3\mathcal{L}_{dist},\end{equation}
where coefficients $\mu_1$, $\mu_2$, and $\mu_3$ are used for balancing between them. 
To facilitate the multi-level design, the final total loss is the weighted average of total losses for different levels as
\begin{equation}\label{eqn:final_total_loss}
    \mathcal{L}=\sum_{i=1}^{n}\lambda_i\mathcal{L}_{i},
\end{equation}
where $n$ is the number of features that are used to compute the loss function, and $\{\lambda_i|i=1, 2, \cdots, n\}$ are coefficients to weight the loss functions of different levels. Normally, the loss functions at later stages should account for more, that is, $\lambda_1<\lambda_2<\cdots<\lambda_n$.
\section{Experiments}\label{chap:experiments}
\subsection{Datasets}
To demonstrate the efficacy of the proposed network, experiments are conducted on three datasets, DFC19, GBH, and ISPRS Vaihingen.
\subsubsection{DFC19}
DFC19 (Data Fusion Contest 19) dataset \cite{bosch2019, christie2020, christie2021, lesaux2019} provides multi-date satellite images and ground truth geometric and semantic labels in Jacksonville, Florida, and Omaha, Nebraska, USA. The images cover around 100 $\text{km}^2$ and date from 2014 to 2016, with a GSD of 1.3 m. The geometric labels are derived from airborne LiDAR data with a nominal pulse spacing of 80 cm. The dataset is delivered as 2783 triplets of images, nDSMs, and semantic maps of size 1024$\times$1024, and GSD of 1.3 m. The semantic maps are processed with only building footprints preserved. To conduct the experiments, the patches are cropped into 44,258 smaller patches of size 256$\times$256 due to GPU memory limits, randomly split into training, validation, and test set, with 31,152, 4432, and 8944 data samples.
\subsubsection{GBH}
Existing datasets, including the DFC19 dataset, lack either amount or diversity, so a new dataset---global building height (GBH) dataset is proposed and used to demonstrate the efficacy of the proposed network. The GBH dataset is composed of optical remote sensing images from PLANET, height maps in the form of nDSMs, and building footprint maps.

The nDSMs are generated by processing open LiDAR point cloud observations from the authorities. First, the point clouds are de-noised, then the height values of all points and the height values of ground points are rasterized into DSMs and digital terrain models (DTMs), respectively. Finally, the normalized height is obtained by simply subtracting DTMs from the corresponding DSMs. Besides, building footprint maps are included in the dataset for testing in this paper. 

The dataset of the current version covers 19 diverse cities around the world and the period from 2013 to 2021. With a patch size of 256$\times$256 and a GSD of 3 m, the dataset is delivered as 20,532 patches, divided into training, validation, and test sets, with 14,971, 3660, and 1901 patches, respectively. Apart from the 19 cities, three cities, Los Angeles, Sao Paulo, and Guangzhou, with 5787 patches, 108 patches, and 1006 patches, respectively, are left out for testing only. It should be noted that only the number of floors for each building is available in Guangzhou, which is converted into building-wise height by assuming a 3 m floor height.
\subsubsection{ISPRS Vaihingen}
ISPRS Vaihingen dataset \cite{isprs,gerke2014} contains aerial orthophotos in IRRG bands, nDSMs generated from LiDAR point clouds, and the corresponding semantic labels, in 33 tiles, with the GSD of 0.09 m. Due to GPU memory limits, they are cropped into patches of size $256\times 256$, randomly split into training, validation, and test set, with 1209, 279, and 248 data samples.
\subsection{Evaluation Metrics}
To evaluate and compare the performances of different models, the predictions are evaluated in terms of RMSE, RMSE-M, RMSE-NM, and RMSE-B. While RMSE is the pixel-wise root mean square error for all pixels, RMSE-M measures the RMSE for only building pixels, and RMSE-NM measures the RMSE for only non-building pixels. To further evaluate the capability of the models to generate LoD-1 building models, the building-wise RMSE, denoted by RMSE-B, is computed with building-wise predicted values and building-wise ground truth values, where the building-wise values are defined as the median of the height values for each building instance represented by one connected component in the building footprint maps.
\subsection{Competitors}
The proposed methods are compared to mainstream FCN-based networks, i.e., SegNet \cite{badrinarayanan2017}, FCN \cite{shelhamer2017}, U-Net \cite{ronneberger2015}, and Efficient U-Net \cite{baheti2020}. Methods in this category are mostly designed for semantic segmentation tasks. To facilitate them for height estimation, the final activation layers for classification, e.g., Sigmoid or Softmax activation, are removed. The output from the last convolutional layer is taken as the predicted height values. Besides, five networks specifically designed for monocular height estimation are tested for comparison, among which IM2HEIGHT \cite{mou2018}, Amirkolaee \textit{et al.} \cite{amirkolaee2019}, and PLNet \cite{xing2021} are FCN networks taking the problem as a regression task; DORN \cite{fu2018,li2020} and Sun \textit{et al.} \cite{sun2022a,li2022} covert the regression task into a classification one, with different bin discretization strategies. The architectures of these networks remain unchanged. 
\subsection{Implementation Details}
The backbone network is built on U-Net \cite{ronneberger2015} and EfficientNet \cite{tan2020} with \cite{alhashim2019} as the decoder, which gives decoded features of five levels, $\{\mathbf{F}_1,\mathbf{F}_2,\mathbf{F}_3,\mathbf{F}_4,\mathbf{F}_5\}$, among which $\{\mathbf{F}_2, \mathbf{F}_3,\mathbf{F}_4,\mathbf{F}_5\}$ are fed into the rest of the network. For the HTC-AdaBins module, the features are divided into patches of size 4, the number of bins is fixed as 256, and 256 tokens are selected to generate foreground and background embeddings, which complies with the output channel number of the convolutional layer in the local branch. For the hybrid regression process, Gaussian distributions are chosen as the reference distributions for foreground pixels, and uniform distributions are assumed for background pixels. For the loss function, the loss components are weighted with the factors. We set $\mu_1=0.01$, $\mu_2=\mu_3=1$, $\lambda_1=0$ ($\mathbf{F}_1$ is discarded), $\lambda_2=0.125$, $\lambda_3=0.25$, $\lambda_4=0.5$, $\lambda_5=1$. The values of the abovementioned hyperparameters are proven to be reasonable through experiments.

The proposed networks are trained with the AdamW optimizer, which is a common optimizer used for training ViTs, while the competitors are trained with the Adam optimizer. For all models, the learning rate is set to 1e-4, and early stopping is applied to avoid overfitting. Practically, if the performance of the network fails to improve for 10 epochs, the training terminates.

For more implementation details, please refer to the released code.
\begin{table}[t]
    \centering
    \caption{Experimental Results of DFC19. RMSE: Pixel-wise root mean square error. RMSE-M: pixel-wise RMSE for building pixels, RMSE-NM: pixel-wise RMSE for non-building pixels. RMSE-B: building-wise RMSE. Blocks from top to bottom: universal FCNs, Networks specifically for MHE, and our proposed networks. Colors: green, the best metric; blue, the second best metric. HTC-DC U-Net: HTC-DC Net backend by U-Net \cite{ronneberger2015}; HTC-DC Net B\{0,5,7\}; HTC-DC Net backed by EfficientNet B\{0,5,7\} \cite{tan2020}.}
    \setlength{\tabcolsep}{6.5pt}
    \begin{tabular}{ccccc}
    \toprule
         Network & RMSE & RMSE-M & RMSE-NM & RMSE-B \\
         \midrule
        SegNet  \cite{badrinarayanan2017} & 7.1397  & 14.6632  & 4.9302  & 5.0540  \\ 
        FCN-32s  \cite{shelhamer2017} & 4.4580  & 7.6650  & 3.6814  & 3.8626  \\
        FCN-16s  \cite{shelhamer2017} & 3.8230  & 7.2371  & 2.9113  & 3.6513  \\
        FCN-8s  \cite{shelhamer2017} & 3.8070  & 7.5295  & 2.7626  & 3.7290  \\
        U-Net  \cite{ronneberger2015} & 2.9776  & 5.7762  & 2.2098  & 3.3759  \\ 
        Eff U-Net B0 \cite{baheti2020} & 3.6316  & 7.6800  & 2.3965  & 3.6774  \\ 
        Eff U-Net B5 \cite{baheti2020} & 3.3220  & 6.5484  & 2.4203  & 3.6403  \\ 
        Eff U-Net B7 \cite{baheti2020} & 3.3248  & 6.8710  & 2.2751  & 3.5638  \\ \midrule
        IM2HEIGHT \cite{mou2018} & 4.8139  & 10.5586  & 2.9693  & 4.1436  \\ 
        Amirkolaee \textit{et al.} \cite{amirkolaee2019} & 2.8709  & 5.7860  & 2.0345  & 3.2337  \\
        DORN  \cite{fu2018,li2020} & 3.5755  & 6.4091  & \color{RoyalBlue}\textbf{1.9402}  & 3.2285  \\ 
        PLNet \cite{xing2021} & 3.2484  & 6.6785  & 2.2397  & 3.5126  \\ 
        Sun \textit{et al.} \cite{li2022,sun2022a} & 5.9381 & 9.3009 & 5.1894 & 5.2313 \\ \midrule
        HTC-DC U-Net & 3.3886 & 7.1287 & 2.2555 & 3.5954 \\
        HTC-DC Net B0 & 2.4894  & 4.3801  & 2.0212  & 3.2484  \\ 
        HTC-DC Net B5 & \color{RoyalBlue}\textbf{2.1813} & \color{RoyalBlue}\textbf{3.8725}  & \color{ForestGreen}\textbf{1.7588}  & \color{RoyalBlue}\textbf{2.7850}  \\ 
        HTC-DC Net B7 & \color{ForestGreen}\textbf{2.1184}  & \color{ForestGreen}\textbf{1.5173}  & 2.1785  & \color{ForestGreen}\textbf{2.3236}  \\ 
         \bottomrule
    \end{tabular}
    \label{tab:res_dfc}
\end{table}
\begin{figure*}[!ht]
    \centering
    \includegraphics[width=\linewidth]{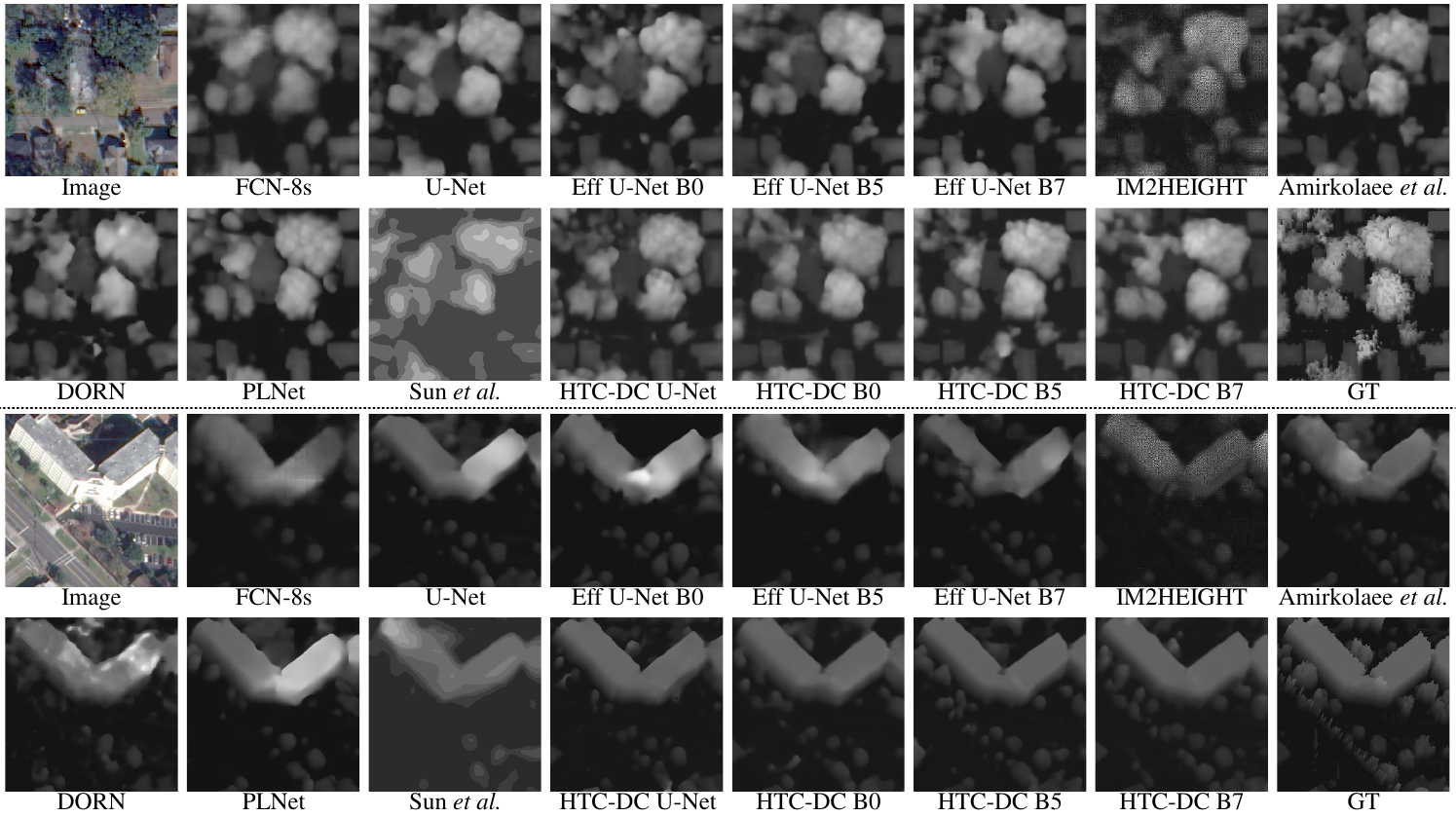}
    \caption{Qualitative results of different models on DFC19. The maps are scaled to the same range.}
    \label{fig:res_vis_dfc}
\end{figure*}
\section{Results}\label{chap:results}
The quantitative results are shown in Table \ref{tab:res_dfc}, Table \ref{tab:res_gbh1}, Table \ref{tab:res_gbh2}, and Table \ref{tab:res_isprs}. The qualitative results are presented in Fig. \ref{fig:res_vis_dfc} and Fig. \ref{fig:res_vis_gbh} for DFC19 and GBH. Generally, our proposed networks gain better results compared to the existing methods, mostly by large margins.
\begin{table}[!t]
    \centering
    \caption{Experimental Results of GBH cities seen during the training.}
    \setlength{\tabcolsep}{6.5pt}
    \begin{tabular}{ccccc}
    \toprule
         Network & RMSE & RMSE-M & RMSE-NM & RMSE-B \\
         \midrule
        SegNet  \cite{badrinarayanan2017} & 7.5876  & 13.7589  & 5.0450  & 8.1552  \\ 
        FCN-32s  \cite{shelhamer2017} & 5.6807  & 8.7639  & 4.6273  & 5.1506  \\ 
        FCN-16s  \cite{shelhamer2017} & 5.2707  & 8.0504  & 4.3304  & 4.7862  \\
        FCN-8s  \cite{shelhamer2017} & 5.0394  & 7.6875  & 4.1447  & 4.3994  \\ 
        U-Net  \cite{ronneberger2015} & 4.5784  & 6.7854  & 3.8532  & 3.7871  \\ 
        Eff U-Net B0 \cite{baheti2020} & 5.4468  & 8.7739  & 4.2581  & 4.8907  \\ 
        Eff U-Net B5 \cite{baheti2020} & 4.9659  & 7.6708  & 4.0406  & 4.3232  \\ 
        Eff U-Net B7 \cite{baheti2020} & 4.8612  & 7.3573  & 4.0244  & 4.1122  \\ \midrule
        IM2HEIGHT \cite{mou2018} & 5.5630  & 9.1883  & 4.2319  & 5.0642  \\ 
        Amirkolaee \textit{et al.} \cite{amirkolaee2019} & 4.9693  & 7.8766  & 3.9482  & 4.4514  \\ 
        DORN  \cite{fu2018,li2020} & 5.4389  & 7.6557  & 4.4233  & 4.3266  \\ 
        PLNet \cite{xing2021} & 4.9334  & 7.7507  & 3.9529  & 4.2653  \\ 
        Sun \textit{et al.} \cite{li2022,sun2022a} & 7.6401  & 10.1139  & 6.9038  & 5.9344 \\ \midrule
        HTC-DC U-Net & 4.5647  & 6.5980  & 3.9119  & \color{RoyalBlue}\textbf{3.5237}  \\
        HTC-DC Net B0 & 4.5200  & 6.6479  & 3.8257  & 3.5321  \\ 
        HTC-DC Net B5 & \color{ForestGreen}\textbf{4.4533}  & \color{ForestGreen}\textbf{6.4433}  & \color{RoyalBlue}\textbf{3.8138}  & \color{ForestGreen}\textbf{3.4391}  \\ 
        HTC-DC Net B7 & \color{RoyalBlue}\textbf{4.4860}  & \color{RoyalBlue}\textbf{6.5848}  & \color{ForestGreen}\textbf{3.8025}  & 3.6158 \\ 
         \bottomrule
    \end{tabular}
    \label{tab:res_gbh1}
\end{table}
\begin{table*}[!t]
    \centering
    \caption{Experimental Results of GBH left-out test cities.}
    \setlength{\tabcolsep}{6.5pt}
    \begin{tabular}{cccccccccccccc}
    \toprule
         \multirow{2}{*}{Network} & \multicolumn{4}{c}{Los Angeles} & \multicolumn{4}{c}{Sao Paulo} & Guangzhou \\
          & RMSE & RMSE-M & RMSE-NM & RMSE-B & RMSE & RMSE-M & RMSE-NM & RMSE-B & RMSE-B \\
         \cmidrule(lr){1-1} \cmidrule(lr){2-5} \cmidrule(lr){6-9} \cmidrule(lr){10-10} 
        SegNet  \cite{badrinarayanan2017} & 4.1720  & 6.4571  & \color{ForestGreen}\textbf{3.2450}  & 3.8522  & 12.2260  & 14.9767  & 9.8720  & 13.4439  & 18.1429  \\ 
        FCN-32s  \cite{shelhamer2017}  & 4.0610  & 4.2778  & 3.9975  & 2.3824  & 10.6806  & 13.2142  & 8.4847  & 11.6005  & 15.8830  \\
        FCN-16s  \cite{shelhamer2017} & 3.7616  & 4.2358  & 3.6161  & 2.4009  & 10.6786  & 13.3031  & 8.3832  & 11.7265  & 15.8745  \\ 
        FCN-8s  \cite{shelhamer2017} & 3.6443  & 4.1717  & 3.4806  & 2.2986  & 10.4170  & 13.0296  & 8.1198  & 11.5194  & 16.2267  \\ 
        U-Net  \cite{ronneberger2015} & 3.5821  & 4.1806  & 3.3935  & 2.2906  & 10.1971  & 12.4664  & 8.2600  & 10.5967  & 14.3027  \\ 
        Eff U-Net B0 \cite{baheti2020} & 3.7321  & 4.5192  & 3.4771  & 2.5061  & 10.6661  & 13.2319  & 8.4344  & 11.5874  & 16.4835  \\ 
        Eff U-Net B5 \cite{baheti2020} & 3.6993  & 4.1933  & 3.5470  & 2.3577  & 10.4470  & 12.9024  & 8.3236  & 11.2397  & 15.2812  \\ 
        Eff U-Net B7 \cite{baheti2020} & 3.4688  & 3.9907  & 3.3061  & 2.2500  & 10.1528  & 12.4475  & 8.1869  & 10.5511  & 14.3158  \\
        \cmidrule(lr){1-1} \cmidrule(lr){2-5} \cmidrule(lr){6-9} \cmidrule(lr){10-10} 
        IM2HEIGHT \cite{mou2018} & 3.7083  & 4.6525  & 3.3937  & 2.5470  & 10.7472  & 13.3169  & 8.5154  & 11.6647  & 16.6797  \\ 
        Amirkolaee \textit{et al.} \cite{amirkolaee2019} & 3.4889  & 4.1646  & \color{RoyalBlue}\textbf{3.2725}  & 2.3188  & 10.6489  & 13.3401  & 8.2777  & 11.6855  & 16.1313  \\ 
        DORN  \cite{fu2018,li2020} & 3.6347  & 4.1038  & 3.4840  & 2.1486  & 11.6384  & 13.4164  & 8.3278  & 11.8950  & 15.5949  \\ 
        PLNet \cite{xing2021} & 3.6135  & 4.1601  & 3.4430  & 2.2698  & 10.7689  & 13.5445  & 8.3100  & 11.9087  & 15.8127  \\ 
        Sun \textit{et al.} \cite{li2022,sun2022a} & 6.2638  & 4.8582  & 6.6077  & 3.8527  & 10.4421  & 12.5467  & 8.6835  & 11.1389  & 14.4799  \\
        \cmidrule(lr){1-1} \cmidrule(lr){2-5} \cmidrule(lr){6-9} \cmidrule(lr){10-10}
        HTC-DC U-Net & \color{ForestGreen}\textbf{3.3568}  & \color{ForestGreen}\textbf{3.6008}  & 3.2844  & \color{ForestGreen}\textbf{1.7963}  & 9.8322  & 12.3838  & 7.5673  & 10.7356  & 12.6199 \\
        HTC-DC Net B0 & 3.5258  & 3.7310  & 3.4655  & 1.9432  & 9.6818  & 11.8199  & 7.8599  & \color{RoyalBlue}\textbf{10.0732}  & \color{ForestGreen}\textbf{11.4203} \\ 
        HTC-DC Net B5 & 3.4586  & 3.7382  & 3.3752  & 1.9458  & \color{RoyalBlue}\textbf{9.5411}  & \color{RoyalBlue}\textbf{11.7565}  & \color{RoyalBlue}\textbf{7.6311}  & 10.2585  & 11.9607  \\ 
        HTC-DC Net B7 & \color{RoyalBlue}\textbf{3.3769} & \color{RoyalBlue}\textbf{3.6131}  & 3.3070  & \color{RoyalBlue}\textbf{1.8286} & \color{ForestGreen}\textbf{9.3007}  & \color{ForestGreen}\textbf{11.4746}  & \color{ForestGreen}\textbf{7.4234}  & \color{ForestGreen}\textbf{9.8424}  & \color{RoyalBlue}\textbf{11.8398} \\ 
         \bottomrule
    \end{tabular}
    \label{tab:res_gbh2}
\end{table*}
\begin{table}[!t]
    \centering
    \caption{Experimental Results of ISPRS Vaihingen.}
    \setlength{\tabcolsep}{6.5pt}
    \begin{tabular}{ccccc}
    \toprule
         Network & RMSE & RMSE-M & RMSE-NM & RMSE-B \\
         \midrule
        SegNet \cite{badrinarayanan2017} & 3.6613  & 5.5393  & 2.7982  & 4.6567   \\ 
        FCN-32s \cite{shelhamer2017} & 2.6990  & 2.2135  & 2.8379  & 2.1119   \\ 
        FCN-16s \cite{shelhamer2017} & 2.7020  & 2.9367  & 2.6219  & 2.6181   \\ 
        FCN-8s \cite{shelhamer2017} & 2.4646  & 3.0572  & 2.2402  & 2.6190   \\ 
        U-Net \cite{ronneberger2015} & 1.4916  & 1.8532  & 1.3545  & 1.6963   \\ 
        Eff U-Net B0 \cite{baheti2020} & 1.9954  & 2.3565  & 1.8640  & 2.2663   \\ 
        Eff U-Net B5 \cite{baheti2020} & 1.8513  & 2.1748  & 1.7341  & 2.0609   \\ 
        Eff U-Net B7 \cite{baheti2020} & 1.8167  & 2.1821  & 1.6819  & 1.9990   \\ \midrule
        IM2HEIGHT \cite{mou2018} & 2.0775  & 2.4483  & 1.9427  & 1.9390   \\ 
        Amirkolaee \textit{et al.} \cite{amirkolaee2019} & 1.7782  & 2.1016  & 1.6605  & 2.0308   \\ 
        DORN  \cite{fu2018,li2020} & 1.8182  & 1.9255  & 1.5416  & 1.8509   \\ 
        PLNet \cite{xing2021} & 1.8324  & 2.1544  & 1.7157  & 1.9622   \\ 
        Sun \textit{et al.} \cite{li2022,sun2022a} & 2.3735  & 2.3491  & 2.3813  & 2.1973 \\ \midrule
        HTC-DC U-Net & \color{RoyalBlue}\textbf{1.3623}  & \color{RoyalBlue}\textbf{1.7410}  & 1.2151  & 1.6264   \\ 
        HTC-DC Net B0 & 1.4286  & 1.8389  & 1.2681  & 1.6547   \\ 
        HTC-DC Net B5 & 1.3842  & 1.8229  & \color{RoyalBlue}\textbf{1.2092}  & \color{RoyalBlue}\textbf{1.6199}   \\ 
        HTC-DC Net B7 & \color{ForestGreen}\textbf{1.3025} & \color{ForestGreen}\textbf{1.6658}  & \color{ForestGreen}\textbf{1.1612}  & \color{ForestGreen}\textbf{1.5419}   \\ 
        \bottomrule
    \end{tabular}
    \label{tab:res_isprs}
\end{table}
\subsection{DFC19}
As shown in Table \ref{tab:res_dfc}, our proposed networks consistently achieve the best metrics on the DFC19 dataset. Among the competitors, Amirkolaee \textit{et al.} \cite{amirkolaee2019} makes the strongest baseline with the best RMSE on all pixels, U-Net \cite{ronneberger2015} performs the best on building pixels, and DORN \cite{li2020} demonstrates superior results on building instances. However, they are still behind the results from our proposed HTC-DC Nets. For instance, HTC-DC Net B7 outperforms them by margins of 0.7525 m, 4.2589 m, and 0.9049 m, respectively. 

In addition to better quantitative results, our HTC-DC Nets exhibit better preserved minor structures and boundaries and more accurate predictions. As shown in the first qualitative result in Fig. \ref{fig:res_vis_dfc}, while other networks predict height maps where the canopy textures are highly blurred, the HTC-DC Nets' predictions are the closest to the ground truth map. In the second example, a tall building is shot with an oblique angle, so the facade is captured in the image. The HTC-DC Nets better distinguish between facade surfaces and roofs, producing consistent elevation maps to input images with sharper building boundaries. Besides, the rooftop of the building should be smooth as seen in the predictions from HTC-DC Nets and the ground truth, while other networks predict different heights for the two parts of the ``L''-shaped building, and the heights are either overestimated or underestimated.
\subsection{GBH}
Generally, larger variabilities are observed on the GBH dataset, which is a more complex and challenging dataset (see Table \ref{tab:res_gbh1}). Despite U-Net performing the best on all metrics among the competitors, it is still inferior to our proposed HTC-DC Nets, such as HTC-DC Net B7, by margins of 0.0924 m, 0.2007 m, 0.0507 m, and 0.1714 m, respectively. Note that Sun \textit{et al.} \cite{sun2022a} fails to deliver reasonable outputs on the GBH dataset with collapsed classification outputs, probably due to the higher complexity of the dataset.

As for the test results on unseen cities (see Table \ref{tab:res_gbh2}), it is expected that the networks' performances will degrade in cities with significant domain shifts from the training cities. Given that the training cities are mostly located in Europe and North America, the performances of Sao Paulo and Guangzhou are remarkably worse. However, the absolute performance losses of HTC-DC Nets are relatively smaller than other networks.

\begin{figure*}[!ht]
    \centering
    \includegraphics[width=\linewidth]{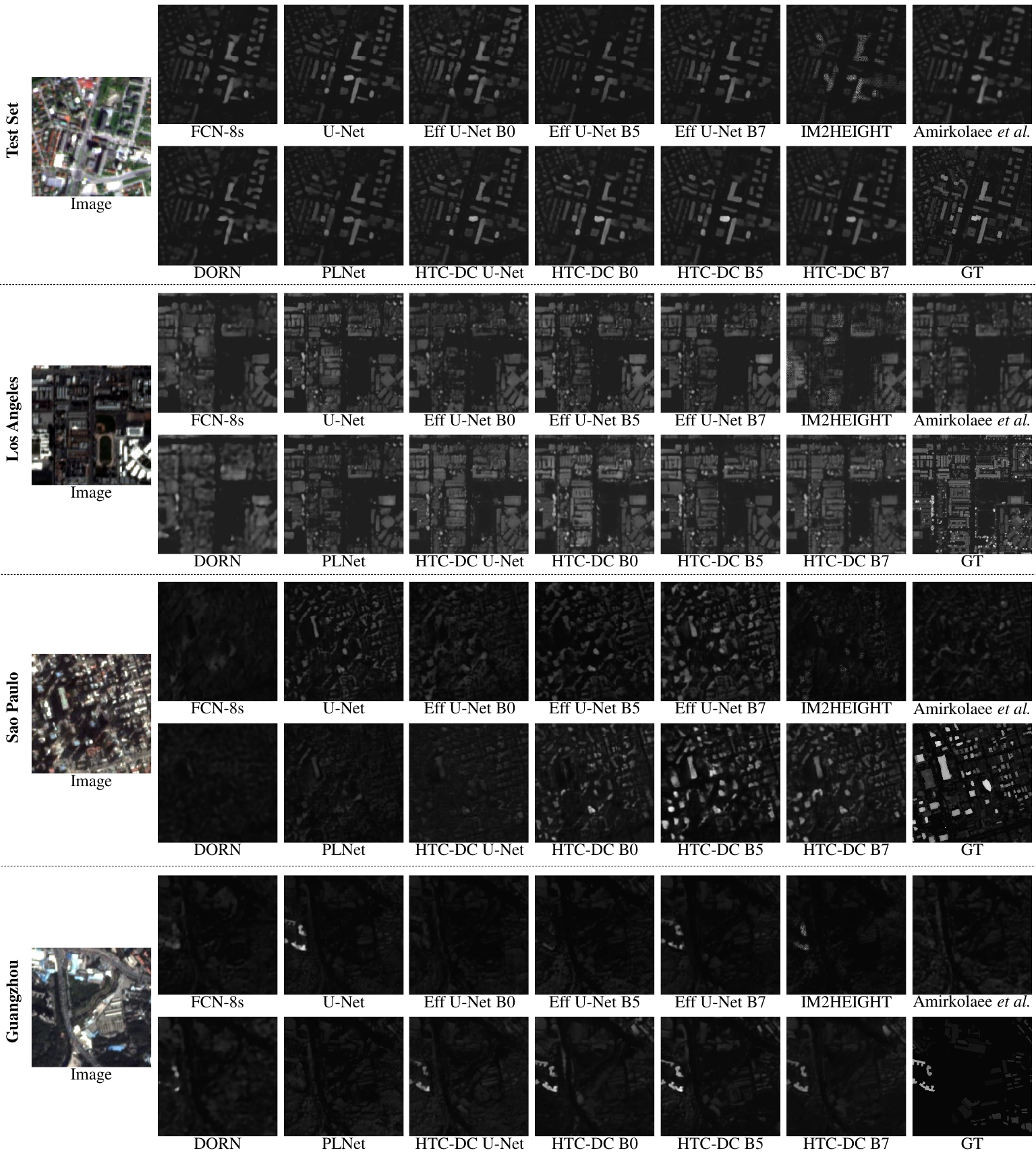}
    \caption{Qualitative results of different models on GBH. The maps are scaled to the same range. Sun \textit{et al.} fails and is, thus, not shown here.}
    \label{fig:res_vis_gbh}
\end{figure*}

In the qualitative results presented in Fig. \ref{fig:res_vis_gbh}, satisfactory outputs are obtained by all the networks on the test set and Los Angeles. However, HTC-DC Nets excel in preserving the minor structures, such as the shapes of the complex buildings. Furthermore, while other networks tend to underestimate the heights of tall buildings, the predicted height value for the tallest building in the first visualization example by HTC-DC Net B5 closely aligns with the ground truth. On Sao Paulo and Guangzhou, other networks show severely degraded performances. For example, in the image sample from Guangzhou, FCNs, Eff U-Nets, IM2HEIGHT, Amirkolaee \textit{et al.}, and PLNet generate height maps where the buildings are almost indiscernible, but our proposed HTC-DC Nets still perform well.
\subsection{ISPRS Vaihingen}
As shown in Table \ref{tab:res_isprs}, our proposed networks outperform the existing methods. The best-performing network, HTC-DC Net B7, surpasses the strongest competitor, U-Net, by margins of 0.1981 m, 0.1874 m, 0.1933 m, 0.1544 m, on RMSE, RMSE-M, RMSE-NM, and RMSE-B, respectively.
\begin{table*}[!t]
    \centering
    \caption{Ablation Study on Multi-level Early Injection.}
    \setlength{\tabcolsep}{6.5pt}
    \begin{tabular}{cccccccccc}
    \toprule
         \multirow{2}{*}{Setting} & \multicolumn{4}{c}{Test Set} & \multicolumn{4}{c}{Los Angeles} & Guangzhou \\
           & RMSE & RMSE-M & RMSE-NM & RMSE-B & RMSE & RMSE-M & RMSE-NM & RMSE-B & RMSE-B \\
         \cmidrule(lr){1-1} \cmidrule(lr){2-5} \cmidrule(lr){6-9} \cmidrule(lr){10-10} 
            $\mathbf{F}_5$ only & 4.5971  & 6.9177  & \color{RoyalBlue}\textbf{3.8233}  & 3.8315  & \color{ForestGreen} \color{ForestGreen}\textbf{3.4492}  & 3.9137  &  \color{ForestGreen}\textbf{3.3058}  & 2.0541  & 12.7593   \\ 
            $\mathbf{F}_2$, $\mathbf{F}_3$, $\mathbf{F}_5$ & 4.6669  & 6.8616  & 3.9511  & 3.7165  & 3.5190  & 3.7368  & 3.4549 &  \color{ForestGreen}\textbf{1.9095}  & 12.2430   \\ 
            $\mathbf{F}_2$, $\mathbf{F}_4$, $\mathbf{F}_5$ & 4.5677  &  \color{ForestGreen}\textbf{6.5485}  & 3.9365  & \color{RoyalBlue}\textbf{3.5469}  & 3.5765  &  \color{ForestGreen}\textbf{3.7216}  & 3.5343  & 1.9639  & \color{RoyalBlue}\textbf{12.1290}   \\ 
            $\mathbf{F}_3$, $\mathbf{F}_4$, $\mathbf{F}_5$ & \color{RoyalBlue}\textbf{4.5207}  & 6.7098  &  \color{ForestGreen}\textbf{3.8003}  & 3.6320  & \color{RoyalBlue}\textbf{3.4832}  & 3.7848  & \color{RoyalBlue}\textbf{3.3930}  & 1.9631  & 12.3482   \\ 
            \color{ForestGreen}\textbf{$\mathbf{F}_2$, $\mathbf{F}_3$, $\mathbf{F}_4$, $\mathbf{F}_5$} &  \color{ForestGreen}\textbf{4.5200}  & \color{RoyalBlue}\textbf{6.6479}  & 3.8257  &  \color{ForestGreen}\textbf{3.5321}  & 3.5258  & \color{RoyalBlue}\textbf{3.7310}  & 3.4655  & \color{RoyalBlue}\textbf{1.9432} &  \color{ForestGreen}\textbf{11.4203}   \\ 
         \bottomrule
    \end{tabular}
    \label{tab:ab_ei}
\end{table*}
\begin{table*}[!h]
    \centering
    \caption{Ablation Study on Head-Tail Cut. HTC: Head-Tail Cut}
    \setlength{\tabcolsep}{1pt}
    \begin{tabular}{cccccccccccccccc}
    \toprule
         \multirow{2}{*}{Setting} & \multicolumn{6}{c}{Test Set} & \multicolumn{6}{c}{Los Angeles} & \multicolumn{2}{c}{Guangzhou} \\
          & Accuracy & RMSE & RMSE-M & RMSE-NM & RMSE-B & RMSE-BG & Accuracy & RMSE & RMSE-M & RMSE-NM & RMSE-B & RMSE-BG & Accuracy & RMSE-B \\
         \cmidrule(lr){1-1} \cmidrule(lr){2-7} \cmidrule(lr){8-13} \cmidrule(lr){14-15}
            w/o HTC & - &  4.5610  & 6.9547  & 3.7526  & 3.8340  & 2.8441  & - & 3.4500  & 3.8475  & 3.3289  & 2.0360  & 2.6987  & - & 12.8314  \\ 
            \color{ForestGreen}\textbf{w/ HTC} & \color{ForestGreen}\textbf{0.7929} & \color{ForestGreen}\textbf{4.5355} & \color{ForestGreen}\textbf{6.8738}  & \color{ForestGreen}\textbf{3.7504}  & \color{ForestGreen}\textbf{3.7047}  & \color{ForestGreen}\textbf{2.7922}  & \color{ForestGreen}\textbf{0.7339} & \color{ForestGreen}\textbf{3.3326}  & \color{ForestGreen}\textbf{3.7958}  & \color{ForestGreen}\textbf{3.1893}  & \color{ForestGreen}\textbf{1.9612}  & \color{ForestGreen}\textbf{2.4739}  & \color{ForestGreen}\textbf{0.5975} & \color{ForestGreen}\textbf{12.4253}  \\ 
         \bottomrule
    \end{tabular}
    \label{tab:ab_htc}
\end{table*}
\begin{figure}[!h]
    \centering
    \includegraphics[width=\linewidth]{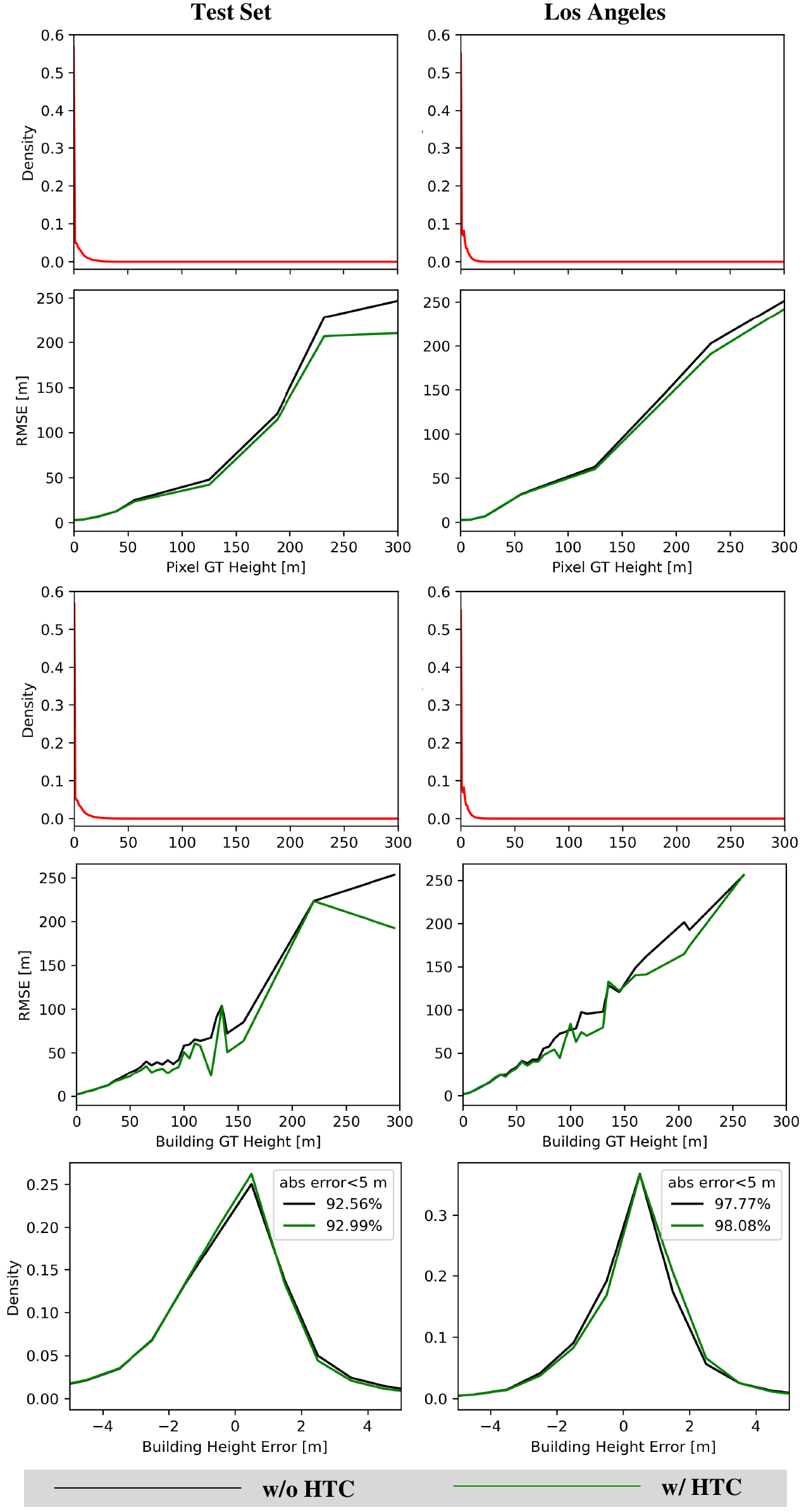}
    \caption{RMSE distribution shows that the head-tail cut helps mitigate the long-tailed effect. From top to bottom: pixel GT height distribution, pixel RMSE vs. pixel GT height, building GT height distribution, building RMSE vs. building GT height, and building height error distribution.}
    \label{fig:ab_htc_rmse_h}
\end{figure}
\\

In general, superior performance is observed on the ISPRS Vaihingen dataset in comparison to the DFC19 dataset, and results on the DFC19 dataset, in turn, are expected to be better than those on the GBH dataset. This observation may be attributed to the resolution differences and the resulting complexity changes between the three datasets. From the experiments, it is concluded that our proposed HTC-DC Nets are advantageous on datasets of various GSDs, namely, the GBH dataset of 3 m GSD, the DFC19 dataset of 1.33 m GSD, and the ISPRS Vaihingen dataset of 0.09 m GSD.
\section{Discussion}\label{chap:discussions}
\subsection{Classification-Regression Paradigm}
The proposed HTC-DC Nets employ the classification-regression paradigm to tackle the monocular height estimation task. The HTC-AdaBins module is responsible for predicting classes and their corresponding probabilities based on the input images. Then, the hybrid regression process combines the predicted classes and class probabilities to obtain the final predictions in the continuous output space.

The classification phase differs from ordinary classification tasks due to the relationship between classes in terms of quantity and their definition varying across different images, which poses great challenges. Consequently, using a simple classification head yields suboptimal results. In the regression phase, the weighted average serves as the smoothing of the related classes. Taking only the values from the classification results for output leads to discrete and unrealistic output maps, as observed in the results from DORN. The need for continuous output maps justifies the introduction of the hybrid regression process.

Previous works have predominantly followed the regression paradigm, where the height values are directly mapped. Besides, several works employing the classification paradigm convert the regression problem into a classification problem, bringing improvements in performance but often coming with manually introduced artifacts. In our experiments, we compare our proposed methods to existing works that follow these paradigms. While almost all the previous works follow the regression paradigm, DORN \cite{li2020} and Sun \textit{et al.} \cite{sun2022a} follow the classification paradigm. Our proposed networks outperform them by significant margins, highlighting the effectiveness of the classification-regression paradigm for monocular height estimation.
\subsection{HTC-AdaBins}
Our proposed HTC-DC Nets are built upon U-Net \cite{ronneberger2015} and EfficientNet \cite{tan2020}. By comparing the results of U-Net, Eff U-Net \cite{baheti2020}, and our proposed HTC-DC Nets, we can demonstrate the efficacy of the HTC-AdaBins module. Notably, our proposed HTC-DC Nets outperform U-Net and Eff U-Nets, particularly for building pixels. This demonstrates that the adaptation to different input images, as well as the incorporation of local and hostile information, addresses the the long-tailed effect, alleviates the underestimation issues, and enhances the performance of monocular height estimation for building pixels.
\subsection{Ablation Studies}
Ablation studies are conducted to show the effectiveness of each design component, with the results reported on the GBH dataset.
\subsubsection{Multi-level Early Injection}
Multi-level features are utilized, so height maps of different scales are predicted for supervision, allowing for supervisory signals to occur earlier in the network. In our proposed HTC-DC Nets, EfficientNet \cite{tan2020} serves as the backbone, and the decoder based on \cite{alhashim2019} generates features maps at five levels, $\{\mathbf{F}_1, \mathbf{F}_2, \mathbf{F}_3, \mathbf{F}_4, \mathbf{F}_5\}$. Typically, the low-level features, such as $\mathbf{F}_1$ are too compact for accurate predictions and are discarded. We select features from $\{\mathbf{F}_2, \mathbf{F}_3, \mathbf{F}_4, \mathbf{F}_5\}$ and report their results.
\begin{figure}[!h]
    \centering
    \includegraphics[width=\linewidth]{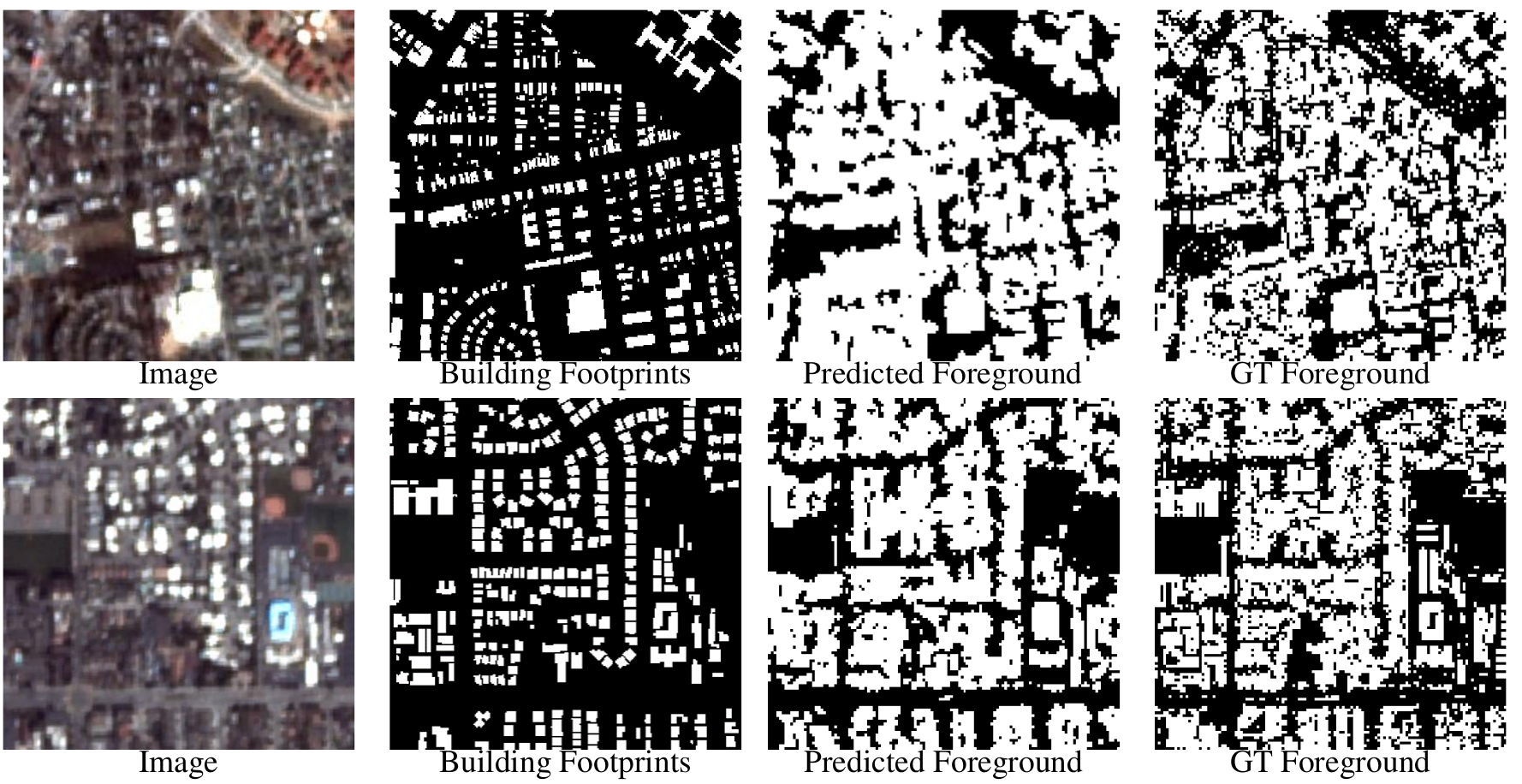}
    \caption{Results from the head-tail cut and the corresponding building footprint masks. The predicted foreground maps are close to the ground truths. For areas with few non-building ground objects (as the second example), the predictions comply with the corresponding building footprint maps.}
    \label{fig:ab_htc}
\end{figure}

The results in Table \ref{tab:ab_ei} show that using features from all four levels yields superior building-related metrics, often ranking among the top two performers. Additionally, it leads to comparable RMSE for all pixels. These findings indicate that fusing features from all stages enhances the networks' performance, especially in improving the accuracy of building height predictions.
\subsubsection{Head-Tail Cut}
To mitigate the negative impact of the majority background pixels on building height predictions, an HTC is employed. Here, we compare the results of networks with and without the HTC. The purpose of the HTC is to improve models' ability to accurately predict building heights, particularly for tall buildings.

Table \ref{tab:ab_htc} illustrates the impact of the HTC on the models' performance. It is obvious that the HTC contributes to the improvement in all metrics. Fig. \ref{fig:ab_htc_rmse_h} presents the distribution of RMSE based on pixel ground truth heights and building ground truth heights. It demonstrates that the models' performance improves significantly for both pixels and buildings with higher values as a result of the HTC. This indicates that the HTC is beneficial for areas where the ground truth heights are higher, leading to improved overall performance. As a consequence, the error distribution is ``squeezed'' toward 0 m, leading to a higher proportion of buildings with absolute errors smaller than 5 m.

Furthermore, regarding the HTC accuracy, since the HTC involves a nearly balanced binary classification task due to the extreme distribution, the classification accuracy is relatively high, as shown in Table \ref{tab:ab_htc}. Additionally, Fig. \ref{fig:ab_htc} presents some visualization results from the HTC, where the predictions are close to the ground truth foreground maps. In areas with few non-building ground objects, such as vegetation, the predicted foreground map accurately corresponds to the corresponding building footprint maps.
\subsubsection{Distribution Constraints as Supervision}\label{sec:ab_dc}
\begin{figure}[!t]
    \centering
    \includegraphics[width=0.5\linewidth]{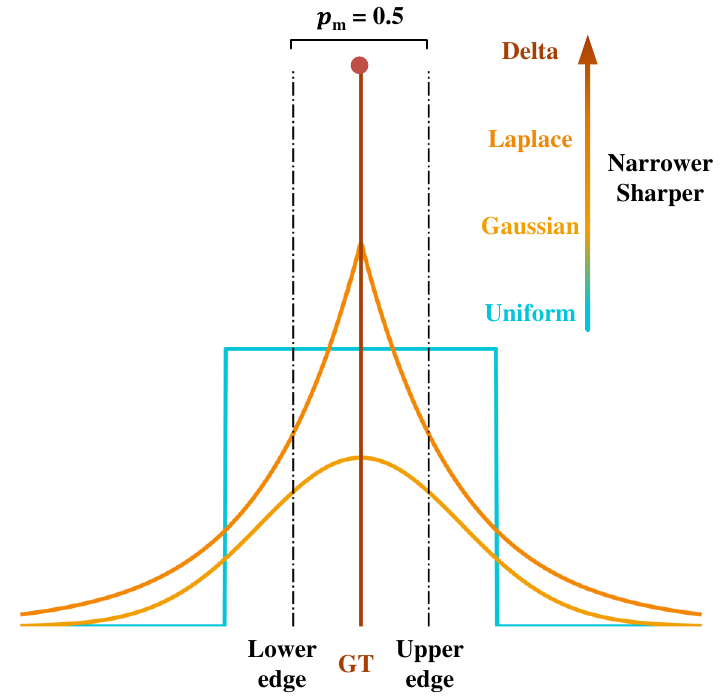}
    \caption{Distributions considered in the ablation study. The same mode probability is assumed for comparison.}
    \label{fig:4distributions}
\end{figure}

From a probabilistic perspective, the weighted average denoted in Eqn. \ref{eqn:hr} is equivalent to computing the expectation value of the underlying height value distribution. Without any constraints, the distribution could be arbitrary, which means the predicted value can often deviate from the mode of the distribution. However, when assuming symmetric unimodal distributions, e.g., Gaussian distribution or Laplace distribution, the predictions should align precisely with the bins where the ground truth values lie. 

The choice of distribution assumption depends on which bins are expected to contribute to the final prediction. If the bins closer to the ground truth bins are supposed to be the primary contributors, then symmetric unimodal distributions are assumed to underlie. In the case where only the ground truth bin is considered for the final prediction, a Delta distribution is assumed. 
\begin{table*}[!h]
    \centering
    \caption{Ablation Study on Distribution-based Constraints for Building Pixels. \# of Improvements: The indicator to show whether a distribution combination works well, defined as the number of metrics where it brings improvements compared to the experiment without using any distribution constraint. The maximum \# of improvements possible in this table is 7.}
    \begin{tabular}{cccccccccc}
    \toprule
         Foreground & Background & \multicolumn{2}{c}{Test Set} & \multicolumn{2}{c}{Los Angeles} & \multicolumn{2}{c}{Sao Paulo} & Guangzhou & \# of\\
         Distribution & Distribution & RMSE-M & RMSE-B & RMSE-M & RMSE-B & RMSE-M & RMSE-B & RMSE-B & Improvements\\
            \cmidrule(lr){1-2} \cmidrule(lr){3-4} \cmidrule(lr){5-6} \cmidrule(lr){7-8} \cmidrule(lr){9-9} \cmidrule(lr){10-10}
            None & None & 6.8738  & 3.7047  & 3.7958  & 1.9612  & 12.2859  & 10.5866  & 12.4253  & - \\ 
            \cmidrule(lr){1-2} \cmidrule(lr){3-4} \cmidrule(lr){5-6} \cmidrule(lr){7-8} \cmidrule(lr){9-9} \cmidrule(lr){10-10}
            \multirow{5}{*}{Uniform} & None & 6.5964  & 3.5745  & 3.7545  & 2.0248  & 11.8237  & 10.1387  & 11.7504  & 6  \\ 
            ~ & Uniform & 6.6901  & 3.5919  & 3.7656  & 1.9524  & 11.8537  & 10.1878  & 12.4098  & \color{ForestGreen}\textbf{7}  \\ 
            ~ & Gaussian & 6.6632  & 3.5927  & 3.6640  & 1.8757  & 11.8343  & 10.2058  & 11.5420  & \color{ForestGreen}\textbf{7}  \\ 
            ~ & Laplace & 6.6833  & 3.5889  & 3.7223  & 1.8896  & 11.9200  & 10.3905  & 12.4644  & 6  \\ 
            ~ & Delta & 6.6525  & 3.5962  & 3.6465  & 1.8986  & 12.0077  & 10.2301  & 12.4980  & 6  \\ 
            \cmidrule(lr){1-2} \cmidrule(lr){3-4} \cmidrule(lr){5-6} \cmidrule(lr){7-8} \cmidrule(lr){9-9} \cmidrule(lr){10-10}
            \multirow{5}{*}{\color{ForestGreen}\textbf{Gaussian}} & None & 6.8569  & 3.6763  & 3.7828  & 1.9475  & 12.1841  & 10.6231  & 13.2616  & 5  \\ 
            ~ & \color{ForestGreen}\textbf{Uniform} & 6.6479  & 3.5321  & 3.7310  & 1.9432  & 11.8199  & 10.0732  & 11.4203  & \color{ForestGreen}\textbf{7}  \\ 
            ~ & Gaussian & 6.6929  & 3.6799  & 3.7553  & 1.9212  & \color{ForestGreen}{\textbf{11.7191}}  & 9.9530  & 11.9169  & \color{ForestGreen}\textbf{7}  \\ 
            ~ & Laplace & 6.9970  & 3.7612  & 3.6898  & \color{ForestGreen}\textbf{1.8303}  & 11.9626  & 10.4068  & 12.9592  & 4  \\ 
            ~ & Delta & 6.7353  & 3.6800  & 3.8820  & 1.9146  & 11.6913  & 10.2209  & 12.5777  & 5  \\ 
            \cmidrule(lr){1-2} \cmidrule(lr){3-4} \cmidrule(lr){5-6} \cmidrule(lr){7-8} \cmidrule(lr){9-9} \cmidrule(lr){10-10}
            \multirow{5}{*}{Laplace} & None & 6.4569  & \color{ForestGreen}\textbf{3.4947}  & 3.6844  & 2.0072  & 11.8162  & \color{ForestGreen}\textbf{9.8691}  & 11.4248  & 6  \\ 
            ~ & Uniform & 7.1464  & 3.8750  & 3.7440  & 1.9064  & 11.8517  & 10.2284  & 12.6783  & 4  \\ 
            ~ & Gaussian & 6.7454  & 3.5648  & 3.6051  & 1.8914  & 11.7454  & 10.1551  & \color{ForestGreen}\textbf{11.1319}  & \color{ForestGreen}\textbf{7}  \\ 
            ~ & Laplace & 6.5990  & 3.6187  & \color{ForestGreen}\textbf{3.5820}  & 1.8972  & 12.1580  & 10.5920  & 12.4754  & 5  \\ 
            ~ & Delta & 6.9813  & 3.9177  & 3.7771  & 1.9472  & 11.8825  & 10.1673  & 13.0317  & 4  \\ 
            \cmidrule(lr){1-2} \cmidrule(lr){3-4} \cmidrule(lr){5-6} \cmidrule(lr){7-8} \cmidrule(lr){9-9} \cmidrule(lr){10-10}
            \multirow{5}{*}{Delta} & None & 6.7221  & 3.6314  & 3.8287  & 2.2092  & 11.8509  & 10.2089  & 11.5422  & 5  \\ 
            ~ & Uniform & 6.7916  & 3.7792  & 3.8050  & 2.1537  & 11.9847  & 10.3700  & 13.2253  & 3  \\ 
            ~ & Gaussian & 6.6732  & 3.6277  & 3.7236  & 2.0684  & 11.8418  & 10.3294  & 12.1994  & 6  \\ 
            ~ & Laplace & 6.8176  & 3.7293  & 3.7665  & 2.1376  & 12.1356  & 10.5050  & 12.1119  & 5  \\ 
            ~ & Delta & \color{ForestGreen}\textbf{6.4508}  & 3.5147  & 3.6891  & 2.0501  & 11.9219  & 10.1786  & 11.8819  & 6 \\ 
     \bottomrule
    \end{tabular}
    \label{tab:ab_probloss_buildings}
\end{table*}
\begin{table*}[!h]
    \centering
    \caption{Ablation Study on Distribution-based Constraints for Non-Building Pixels. \# of Improvements: The indicator to show whether a distribution combination works well, defined as the number of metrics where it brings improvements compared to the experiment without using any distribution constraint. The maximum \# of improvements possible in this table is 6.}
    \begin{tabular}{ccccccccc}
    \toprule
         Foreground & Background & \multicolumn{2}{c}{Test Set} & \multicolumn{2}{c}{Los Angeles} & \multicolumn{2}{c}{Sao Paulo} & \# of\\
         Distribution & Distribution & RMSE-NM & RMSE-BG & RMSE-NM & RMSE-BG & RMSE-NM & RMSE-BG & Improvements\\
        \cmidrule(lr){1-2} \cmidrule(lr){3-4} \cmidrule(lr){5-6} \cmidrule(lr){7-8} \cmidrule(lr){9-9}
        \color{ForestGreen}\textbf{None} & \color{ForestGreen}\textbf{None} & 3.7504  & \color{ForestGreen}\textbf{2.7922}  & \color{ForestGreen}\textbf{3.1893}  & \color{ForestGreen}\textbf{2.4739}  & 7.8393  & 4.7616  & - \\ 
        \cmidrule(lr){1-2} \cmidrule(lr){3-4} \cmidrule(lr){5-6} \cmidrule(lr){7-8} \cmidrule(lr){9-9}
        \multirow{5}{*}{Uniform} & None & 3.8793  & 3.2119  & 3.5237  & 3.0009  & 7.7487  & 5.6451  & 1   \\ 
        ~ & Uniform & 3.8353  & 3.1790  & 3.3873  & 2.8101  & \color{ForestGreen}\textbf{7.5350}  & 4.8217  & 1   \\ 
        ~ & Gaussian & 3.8616  & 3.2373  & 3.4049  & 2.9400  & 7.8226  & 5.7859  & 1   \\ 
        ~ & Laplace & 3.9288  & 3.3928  & 3.5136  & 3.1256  & 7.6822  & 5.0013  & 1   \\ 
        ~ & Delta & 3.9545  & 3.4338  & 3.5413  & 3.1895  & 7.9198  & 5.6409  & 0   \\ 
        \cmidrule(lr){1-2} \cmidrule(lr){3-4} \cmidrule(lr){5-6} \cmidrule(lr){7-8} \cmidrule(lr){9-9}
        \multirow{5}{*}{Gaussian} & None & 3.8072  & 3.0431  & 3.3036  & 2.6715  & 7.7254  & \color{ForestGreen}\textbf{4.4736}  & \color{ForestGreen}\textbf{2}   \\ 
        ~ & Uniform & 3.8257  & 3.1978  & 3.4655  & 2.9607  & 7.8599  & 5.8695  & 0   \\ 
        ~ & Gaussian & 3.8501  & 3.1883  & 3.3374  & 2.8094  & 7.8222  & 6.1055  & 1   \\ 
        ~ & Laplace & 3.8537  & 3.2602  & 3.4332  & 3.0718  & 7.7656  & 5.8894  & 1   \\ 
        ~ & Delta & 4.0374  & 3.5799  & 3.6969  & 3.4024  & 7.6865  & 5.5319  & 1   \\ 
        \cmidrule(lr){1-2} \cmidrule(lr){3-4} \cmidrule(lr){5-6} \cmidrule(lr){7-8} \cmidrule(lr){9-9}
        \multirow{5}{*}{Laplace} & None & 4.0553  & 3.5428  & 3.7021  & 3.2965  & 8.0771  & 6.6759  & 0   \\ 
        ~ & Uniform & \color{ForestGreen}\textbf{3.7403}  & 3.0639  & 3.4464  & 3.0200  & 7.5878  & 5.1780  & \color{ForestGreen}\textbf{2}   \\ 
        ~ & Gaussian & 3.8631  & 3.3418  & 3.6327  & 3.3170  & 7.7828  & 5.9134  & 1   \\ 
        ~ & Laplace & 4.1204  & 3.8706  & 3.9178  & 3.8048  & 7.6909  & 5.3824  & 1   \\ 
        ~ & Delta & 3.9372  & 3.4197  & 3.5051  & 3.1671  & 7.8802  & 5.7204  & 0   \\ 
        \cmidrule(lr){1-2} \cmidrule(lr){3-4} \cmidrule(lr){5-6} \cmidrule(lr){7-8} \cmidrule(lr){9-9}
        \multirow{5}{*}{Delta} & None & 3.9709  & 3.3611  & 3.7854  & 3.3787  & 7.7560  & 5.4105  & 1   \\ 
        ~ & Uniform & 4.0210  & 3.5332  & 3.7792  & 3.3955  & 7.6475  & 5.1138  & 1   \\ 
        ~ & Gaussian & 4.0375  & 3.6006  & 3.7515  & 3.4089  & 7.7033  & 5.5643  & 1   \\ 
        ~ & Laplace & 3.9850  & 3.5221  & 3.7192  & 3.3321  & 7.8084  & 5.5763  & 1   \\ 
        ~ & Delta & 4.1970  & 3.9896  & 4.0407  & 3.9340  & 7.9399  & 6.1491  & 0   \\ 
     \bottomrule
    \end{tabular}
    \label{tab:ab_probloss_nonbuilding}
\end{table*}
\begin{table*}[!h]
    \centering
    \caption{Ablation Study on Distribution-based Constraints for All Pixels. RMSEs for all pixels are shown. \# of Improvements: The indicator to show whether a distribution combination works well, defined as the number of metrics where it brings improvements compared to the experiment without using any distribution constraint. The maximum \# of improvements possible in this table is 3.}
    \begin{tabular}{ccccccc}
    \toprule
         Foreground Distribution & Background Distribution & Test Set & Los Angeles & Sao Paulo & \# of Improvements\\
         \midrule
            None & None & 4.5355  & \color{ForestGreen}\textbf{3.3326}  & 9.9072  & - \\ 
            \midrule
            \multirow{5}{*}{Uniform} & None & 4.5418  & 3.5759  & 9.6307  & 1  \\ 
            ~ & Uniform & 4.5386  & 3.4743  & 9.5454  & 1  \\ 
            ~ & Gaussian & 4.5488  & 3.4637  & 9.6712  & 1  \\ 
            ~ & Laplace & 4.6006  & 3.5607  & 9.6479  & 1  \\ 
            ~ & Delta & 4.6096  & 3.5648  & 9.8046  & 1  \\ 
            \midrule
            \multirow{5}{*}{\color{ForestGreen}\textbf{Gaussian}} & None & 4.5684  & 3.4151  & 9.8023  & 1  \\ 
            ~ & \color{ForestGreen}\textbf{Uniform} & \color{ForestGreen}\textbf{4.5200}  & 3.5258  & 9.6818  & \color{ForestGreen}\textbf{2}  \\ 
            ~ & Gaussian & 4.5496  & 3.4340  & 9.6133  & 1  \\ 
            ~ & Laplace & 4.6408  & 3.4915  & 9.7087  & 1  \\ 
            ~ & Delta & 4.6902  & 3.7385  & \color{ForestGreen}\textbf{9.5345}  & 1  \\ 
            \midrule
            \multirow{5}{*}{Laplace} & None & 4.6254  & 3.6982  & 9.7848  & 1  \\ 
            ~ & Uniform & 4.6107  & 3.5142  & 9.5690  & 1  \\ 
            ~ & Gaussian & 4.5736  & 3.6266  & 9.6076  & 1  \\ 
            ~ & Laplace & 4.7102  & 3.8462  & 9.7730  & 1  \\ 
            ~ & Delta & 4.6922  & 3.5669  & 9.7229  & 1  \\ 
            \midrule
            \multirow{5}{*}{Delta} & None & 4.6405  & 3.7950  & 9.6478  & 1  \\ 
            ~ & Uniform & 4.6947  & 3.7849  & 9.6645  & 1  \\ 
            ~ & Gaussian & 4.6728  & 3.7454  & 9.6183  & 1  \\ 
            ~ & Laplace & 4.6774  & 3.7297  & 9.8165  & 1  \\ 
            ~ & Delta & 4.7243  & 3.9658  & 9.7712  & 1 \\ 
     \bottomrule
    \end{tabular}
    \label{tab:ab_probloss_overall}
\end{table*}
\begin{figure*}[!h]
    \centering
    \includegraphics[width=\linewidth]{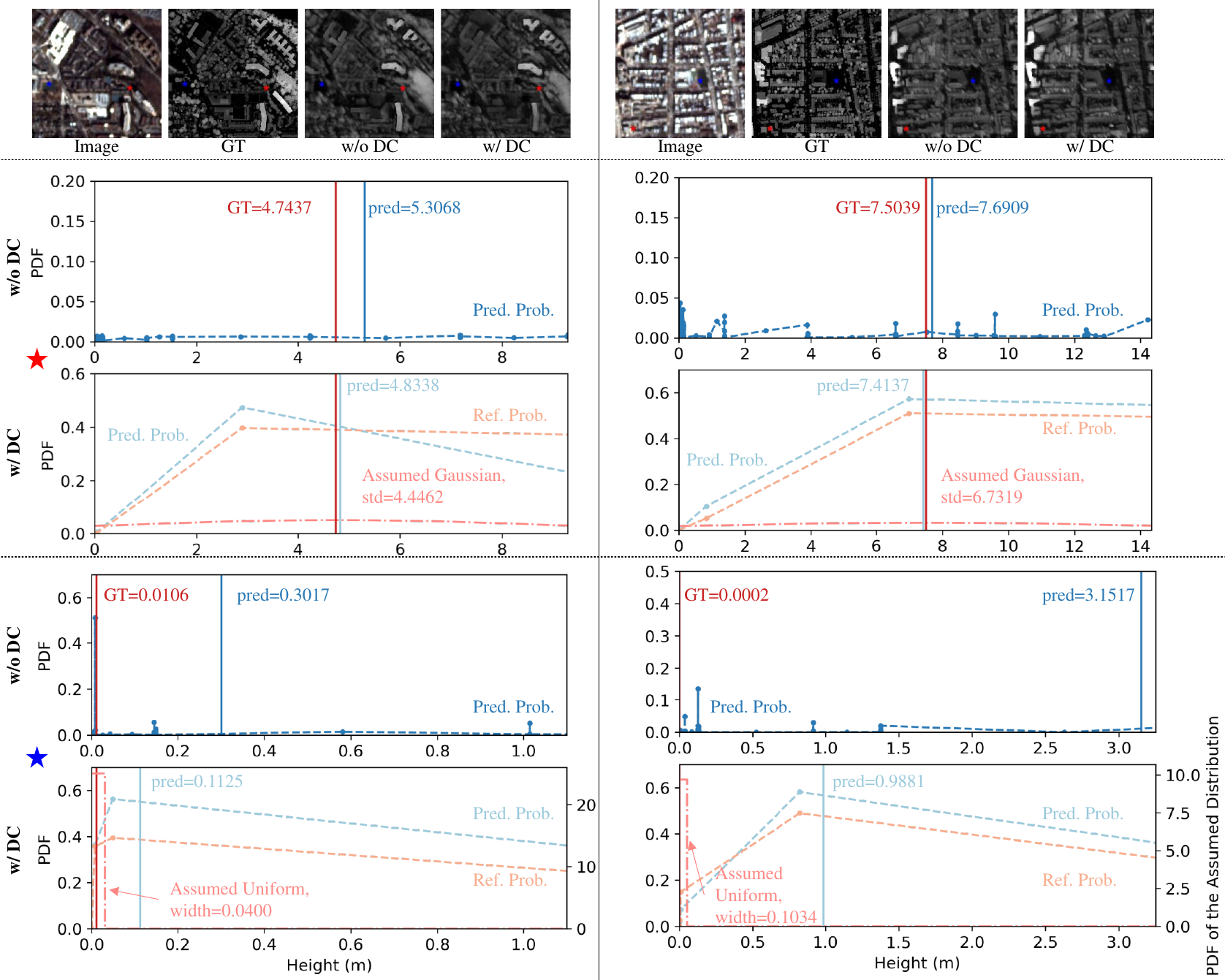}
    \caption{Visualization of the ablation study for the distribution-based constraints (DCs). Two pixels from the predicted foreground (FG) and the background (BG) are taken for comparison, which are annotated by the red and blue stars, respectively. With the distribution-based constraints, the predicted bin probabilities are regularized, so the predicted height values are more accurate.}
    \label{fig:ab_probloss}
\end{figure*}

The choice of distribution also determines the extent to which the bins contribute. Within the family of symmetric unimodal distributions, the main difference lies in the sharpness of the peaks, which represents the margin between the mode probability and the probabilities of the surrounding bins. If all supporting bins are expected to contribute equally, a uniform distribution is assumed.

In the ablation study, four different distributions are implemented to demonstrate the optimality of using Gaussian distributions for the foreground and uniform distributions for the background, in terms of selecting bin contributors and determining their contribution amount. The four distributions used are Gaussian, Laplace, Delta, and uniform distributions. (refer to Fig. \ref{fig:4distributions}). Among these, the Delta distribution has the sharpest and narrowest peak, while the uniform distribution has the smoothest and broadest peak. As a complement to Section \ref{sec:dbc}, the equation for the derivation of the scale parameter from the mode probability of a Laplace distribution $h\sim\mathcal{L}(h|\tilde{h},b)=\displaystyle\frac{1}{2b}\exp(-\frac{|h-\tilde{h}|}{b})$ is provided without proof as
\begin{equation}
     b=\displaystyle-\frac{e_{m+1}-e_m}{2\ln(1-P_m)}.
\end{equation}

As there is always a compromise between the prediction for building pixels and non-building pixels, the experiments are evaluated in three folds. The results are presented in Table \ref{tab:ab_probloss_buildings} for building pixels, Table \ref{tab:ab_probloss_nonbuilding} for non-building pixels, and Table \ref{tab:ab_probloss_overall} for all pixels. The indicator of a well-performing distribution is its ability to consistently bring improvements compared to experiments without any distribution constraint, which is represented by the column ``\# of Improvements'' in each table. Based on the results, the combination of Gaussian for foreground and uniform for background yields the largest number of improvements on building pixels (7), and all pixels (2). Considering the buildings are of greater interest, this combination is selected as the final configuration. We argue that the choices of distributions have a great effect on the final performances; however, it is hard to analytically decide which distributions to use.

Fig. \ref{fig:ab_probloss} visualizes the predicted bin probabilities from networks without and with the DC. Visually, the application of the distribution constraints has a subtle effect on the predicted height maps. However, when examining the bin probability graphs, it is evident that without constraints, the predicted probabilities are relatively small and disorganized, and the predicted values result from a wider range of bin centers. After the constraints are applied, the predicted probabilities are pushed toward the reference bin probabilities derived from the assumed underlying distribution. This indicates that the constraints effectively regularize the bin probabilities and align them with the bins near the ground truth values, as assumed. The bin probability patterns with constraints result in improvements to the hybrid regression results.

It is important to note that there are some failure cases when the mode probabilities are relatively small, leading to extremely large derived standard deviations. This causes the assumed distributions to approach uniform distributions. This phenomenon is likely due to domain shifts, as these failure cases occur more frequently in certain cities.
\section{Conclusion}\label{chap:conclusions}
We propose HTC-DC Net, a network for predicting heights from single remote sensing images. The proposed network utilizes a classification-regression paradigm with a ViT to incorporate holistic features and local features. The regression phase with hybrid regression acts as a smoothing process for the classification phase conducted by the HTC-AdaBins module. With the DCs, the height predictions are efficiently regularized. Besides, to combat the long-tailed distribution problems, a novel HTC is conducted to separate the foregrounds from the backgrounds for different treatments. Experiments show that our proposed HTC-DC Net achieves state-of-the-art performance.

Despite the impressive results given by our proposed HTC-DC Net, the domain shifts between different cities are still challenging for large-scale applications. They lead to performance drops, especially for cities with distinct urban morphologies. Therefore, further works could be done to address the domain shifts by applying domain generalization techniques.

\section*{Acknowledgment}
This work was supported by the Helmholtz Association's Initiative and Networking Fund on the HAICORE@KIT partition and the HAICORE@FZJ partition. The authors would like to also thank Y. Cao for easier access to building height data in China, and S. Xing for providing the code of PLNet \cite{xing2021}.

\ifCLASSOPTIONcaptionsoff
  \newpage
\fi

\bibliographystyle{IEEEtran}
\bibliography{refs}

\begin{thebibliography}{10}
\providecommand{\url}[1]{#1}
\csname url@samestyle\endcsname
\providecommand{\newblock}{\relax}
\providecommand{\bibinfo}[2]{#2}
\providecommand{\BIBentrySTDinterwordspacing}{\spaceskip=0pt\relax}
\providecommand{\BIBentryALTinterwordstretchfactor}{4}
\providecommand{\BIBentryALTinterwordspacing}{\spaceskip=\fontdimen2\font plus
\BIBentryALTinterwordstretchfactor\fontdimen3\font minus
  \fontdimen4\font\relax}
\providecommand{\BIBforeignlanguage}[2]{{%
\expandafter\ifx\csname l@#1\endcsname\relax
\typeout{** WARNING: IEEEtran.bst: No hyphenation pattern has been}%
\typeout{** loaded for the language `#1'. Using the pattern for}%
\typeout{** the default language instead.}%
\else
\language=\csname l@#1\endcsname
\fi
#2}}
\providecommand{\BIBdecl}{\relax}
\BIBdecl

\bibitem{arefi2013}
H.~Arefi and P.~Reinartz, ``Building {{Reconstruction Using DSM}} and
  {{Orthorectified Images}},'' \emph{Remote Sensing}, vol.~5, no.~4, pp.
  1681--1703, Apr. 2013.

\bibitem{partovi2014}
T.~Partovi, T.~Krau{\ss}, H.~Arefi, M.~Omidalizarandi, and P.~Reinartz,
  ``Model-driven {{3D}} building reconstruction based on integeration of
  {{DSM}} and spectral information of satellite images,'' in \emph{2014 {{IEEE
  Geoscience}} and {{Remote Sensing Symposium}}}, Jul. 2014, pp. 3168--3171.

\bibitem{wang2021}
Y.~Wang, S.~Zorzi, and K.~Bittner, ``Machine-{{Learned 3D Building
  Vectorization From Satellite Imagery}},'' in \emph{Proceedings of the
  {{IEEE}}/{{CVF Conference}} on {{Computer Vision}} and {{Pattern
  Recognition}}}, 2021, pp. 1072--1081.

\bibitem{matese2017}
A.~Matese, S.~F. Di~Gennaro, and A.~Berton, ``Assessment of a canopy height
  model ({{CHM}}) in a vineyard using {{UAV-based}} multispectral imaging,''
  \emph{International Journal of Remote Sensing}, vol.~38, no. 8-10, pp.
  2150--2160, May 2017.

\bibitem{ota2015}
T.~Ota, M.~Ogawa, K.~Shimizu, T.~Kajisa, N.~Mizoue, S.~Yoshida, G.~Takao,
  Y.~Hirata, N.~Furuya, T.~Sano, H.~Sokh, V.~Ma, E.~Ito, J.~Toriyama, Y.~Monda,
  H.~Saito, Y.~Kiyono, S.~Chann, and N.~Ket, ``Aboveground {{Biomass Estimation
  Using Structure}} from {{Motion Approach}} with {{Aerial Photographs}} in a
  {{Seasonal Tropical Forest}},'' \emph{Forests}, vol.~6, no.~11, pp.
  3882--3898, Nov. 2015.

\bibitem{sadeghi2016}
Y.~Sadeghi, B.~{St-Onge}, B.~Leblon, and M.~Simard, ``Canopy {{Height Model}}
  ({{CHM}}) {{Derived From}} a {{TanDEM-X InSAR DSM}} and an {{Airborne Lidar
  DTM}} in {{Boreal Forest}},'' \emph{IEEE Journal of Selected Topics in
  Applied Earth Observations and Remote Sensing}, vol.~9, no.~1, pp. 381--397,
  Jan. 2016.

\bibitem{priestnall2000}
G.~Priestnall, J.~Jaafar, and A.~Duncan, ``Extracting urban features from
  {{LiDAR}} digital surface models,'' \emph{Computers, Environment and Urban
  Systems}, vol.~24, no.~2, pp. 65--78, Mar. 2000.

\bibitem{elaksher}
A.~F. Elaksher, J.~S. Bethel \emph{et~al.}, ``Reconstructing 3d buildings from
  lidar data,'' \emph{International Archives Of Photogrammetry Remote Sensing
  and Spatial Information Sciences}, vol.~34, no. 3/A, pp. 102--107, 2002.

\bibitem{chen2022}
Z.~Chen, H.~Ledoux, S.~Khademi, and L.~Nan, ``Reconstructing compact building
  models from point clouds using deep implicit fields,'' \emph{ISPRS Journal of
  Photogrammetry and Remote Sensing}, vol. 194, pp. 58--73, Dec. 2022.

\bibitem{zhu2014}
X.~X. Zhu and R.~Bamler, ``Superresolving {{SAR Tomography}} for
  {{Multidimensional Imaging}} of {{Urban Areas}}: {{Compressive}}
  sensing-based {{TomoSAR}} inversion,'' \emph{IEEE Signal Processing
  Magazine}, vol.~31, no.~4, pp. 51--58, Jul. 2014.

\bibitem{zhu2014a}
X.~X. Zhu and M.~Shahzad, ``Facade {{Reconstruction Using Multiview Spaceborne
  TomoSAR Point Clouds}},'' \emph{IEEE Transactions on Geoscience and Remote
  Sensing}, vol.~52, no.~6, pp. 3541--3552, Jun. 2014.

\bibitem{sun2022b}
Y.~Sun, Y.~Hua, L.~Mou, and X.~X. Zhu, ``{{CG-Net}}: {{Conditional GIS-Aware
  Network}} for {{Individual Building Segmentation}} in {{VHR SAR Images}},''
  \emph{IEEE Transactions on Geoscience and Remote Sensing}, vol.~60, pp.
  1--15, 2022.

\bibitem{sun2022c}
Y.~Sun, L.~Mou, Y.~Wang, S.~Montazeri, and X.~X. Zhu, ``Large-scale building
  height retrieval from single {{SAR}} imagery based on bounding box regression
  networks,'' \emph{ISPRS Journal of Photogrammetry and Remote Sensing}, vol.
  184, pp. 79--95, Feb. 2022.

\bibitem{shi2020}
Y.~Shi, R.~Bamler, Y.~Wang, and X.~X. Zhu, ``{{SAR Tomography}} at the
  {{Limit}}: {{Building Height Reconstruction Using Only}} 3-5 {{TanDEM-X
  Bistatic Interferograms}},'' \emph{IEEE Transactions on Geoscience and Remote
  Sensing}, vol.~58, no.~11, pp. 8026--8037, Nov. 2020.

\bibitem{tack2012}
F.~Tack, G.~Buyuksalih, and R.~Goossens, ``{{3D}} building reconstruction based
  on given ground plan information and surface models extracted from spaceborne
  imagery,'' \emph{ISPRS Journal of Photogrammetry and Remote Sensing},
  vol.~67, pp. 52--64, Jan. 2012.

\bibitem{ginzler2015}
C.~Ginzler and M.~L. Hobi, ``Countrywide {{Stereo-Image Matching}} for
  {{Updating Digital Surface Models}} in the {{Framework}} of the {{Swiss
  National Forest Inventory}},'' \emph{Remote Sensing}, vol.~7, no.~4, pp.
  4343--4370, Apr. 2015.

\bibitem{xiong2021benchmark}
Z.~Xiong, W.~Huang, J.~Hu, Y.~Shi, Q.~Wang, and X.~X. Zhu, ``{THE benchmark}:
  Transferable representation learning for monocular height estimation,''
  \emph{arXiv preprint arXiv:2112.14985}, 2021.

\bibitem{xiong2022}
Z.~Xiong, F.~Zhang, Y.~Wang, Y.~Shi, and X.~X. Zhu, ``{{EarthNets}}:
  {{Empowering AI}} in {{Earth Observation}},'' Oct. 2022.

\bibitem{huang2007}
X.~Huang and L.~K. Kwoh, ``{{3D}} building reconstruction and visualization for
  single high resolution satellite image,'' in \emph{2007 {{IEEE International
  Geoscience}} and {{Remote Sensing Symposium}}}, Jul. 2007, pp. 5009--5012.

\bibitem{izadi2012}
M.~Izadi and P.~Saeedi, ``Three-{{Dimensional Polygonal Building Model
  Estimation From Single Satellite Images}},'' \emph{IEEE Transactions on
  Geoscience and Remote Sensing}, vol.~50, no.~6, pp. 2254--2272, Jun. 2012.

\bibitem{zhao2020}
C.~Zhao, Q.~Sun, C.~Zhang, Y.~Tang, and F.~Qian, ``Monocular depth estimation
  based on deep learning: {{An}} overview,'' \emph{Science China Technological
  Sciences}, vol.~63, no.~9, pp. 1612--1627, Sep. 2020.

\bibitem{dosovitskiy2021}
A.~Dosovitskiy, L.~Beyer, A.~Kolesnikov, D.~Weissenborn, X.~Zhai,
  T.~Unterthiner, M.~Dehghani, M.~Minderer, G.~Heigold, S.~Gelly, J.~Uszkoreit,
  and N.~Houlsby, ``An {{Image}} is {{Worth}} 16x16 {{Words}}: {{Transformers}}
  for {{Image Recognition}} at {{Scale}},'' Jun. 2021.

\bibitem{bhat2021}
S.~F. Bhat, I.~Alhashim, and P.~Wonka, ``{{AdaBins}}: {{Depth Estimation}}
  using {{Adaptive Bins}},'' in \emph{2021 {{IEEE}}/{{CVF Conference}} on
  {{Computer Vision}} and {{Pattern Recognition}} ({{CVPR}})}, Jun. 2021, pp.
  4008--4017.

\bibitem{zhang2021}
Y.~Zhang, B.~Kang, B.~Hooi, S.~Yan, and J.~Feng, ``Deep long-tailed learning:
  {{A}} survey,'' \emph{arXiv:2110.04596}, 2021.

\bibitem{shelhamer2017}
E.~Shelhamer, J.~Long, and T.~Darrell, ``Fully {{Convolutional Networks}} for
  {{Semantic Segmentation}},'' \emph{IEEE Transactions on Pattern Analysis and
  Machine Intelligence}, vol.~39, no.~4, pp. 640--651, Apr. 2017.

\bibitem{badrinarayanan2017}
V.~Badrinarayanan, A.~Kendall, and R.~Cipolla, ``{{SegNet}}: {{A Deep
  Convolutional Encoder-Decoder Architecture}} for {{Image Segmentation}},''
  \emph{IEEE Transactions on Pattern Analysis and Machine Intelligence},
  vol.~39, no.~12, pp. 2481--2495, Dec. 2017.

\bibitem{ronneberger2015}
O.~Ronneberger, P.~Fischer, and T.~Brox, ``U-{{Net}}: {{Convolutional
  Networks}} for {{Biomedical Image Segmentation}},'' in \emph{Medical {{Image
  Computing}} and {{Computer-Assisted Intervention}} \textendash{} {{MICCAI}}
  2015}, ser. Lecture {{Notes}} in {{Computer Science}}, N.~Navab,
  J.~Hornegger, W.~M. Wells, and A.~F. Frangi, Eds.\hskip 1em plus 0.5em minus
  0.4em\relax {Cham}: {Springer International Publishing}, 2015, pp. 234--241.

\bibitem{baheti2020}
B.~Baheti, S.~Innani, S.~Gajre, and S.~Talbar, ``Eff-{{UNet}}: {{A Novel
  Architecture}} for {{Semantic Segmentation}} in {{Unstructured
  Environment}},'' in \emph{2020 {{IEEE}}/{{CVF Conference}} on {{Computer
  Vision}} and {{Pattern Recognition Workshops}} ({{CVPRW}})}.\hskip 1em plus
  0.5em minus 0.4em\relax {Seattle, WA, USA}: {IEEE}, Jun. 2020, pp.
  1473--1481.

\bibitem{mou2018}
L.~Mou and X.~X. Zhu, ``{{IM2HEIGHT}}: {{Height Estimation}} from {{Single
  Monocular Imagery}} via {{Fully Residual Convolutional-Deconvolutional
  Network}},'' \emph{arXiv:1802.10249}, Feb. 2018.

\bibitem{amirkolaee2019}
H.~A. Amirkolaee and H.~Arefi, ``Height estimation from single aerial images
  using a deep convolutional encoder-decoder network,'' \emph{ISPRS Journal of
  Photogrammetry and Remote Sensing}, vol. 149, pp. 50--66, Mar. 2019.

\bibitem{laina2016}
I.~Laina, C.~Rupprecht, V.~Belagiannis, F.~Tombari, and N.~Navab, ``Deeper
  {{Depth Prediction}} with {{Fully Convolutional Residual Networks}},''
  \emph{arXiv:1606.00373 [cs]}, Sep. 2016.

\bibitem{xing2021}
S.~Xing, Q.~Dong, and Z.~Hu, ``Gated {{Feature Aggregation}} for {{Height
  Estimation From Single Aerial Images}},'' \emph{IEEE Geoscience and Remote
  Sensing Letters}, pp. 1--5, 2021.

\bibitem{xiong2022a}
Z.~Xiong, S.~Chen, Y.~Shi, and X.~X. Zhu, ``Disentangled latent transformer for
  interpretable monocular height estimation,'' \emph{arXiv preprint
  arXiv:2201.06357}, 2022.

\bibitem{srivastava2017}
S.~Srivastava, M.~Volpi, and D.~Tuia, ``Joint height estimation and semantic
  labeling of monocular aerial images with {{CNNS}},'' in \emph{2017 {{IEEE
  International Geoscience}} and {{Remote Sensing Symposium}} ({{IGARSS}})},
  Jul. 2017, pp. 5173--5176.

\bibitem{carvalho2020}
M.~Carvalho, B.~Le~Saux, P.~{Trouve-Peloux}, F.~Champagnat, and A.~Almansa,
  ``Multitask {{Learning}} of {{Height}} and {{Semantics From Aerial
  Images}},'' \emph{IEEE Geoscience and Remote Sensing Letters}, vol.~17,
  no.~8, pp. 1391--1395, Aug. 2020.

\bibitem{elhousni2021}
M.~Elhousni, Z.~Zhang, and X.~Huang, ``Height {{Prediction}} and {{Refinement
  From Aerial Images With Semantic}} and {{Geometric Guidance}},'' \emph{IEEE
  Access}, vol.~9, pp. 145\,638--145\,647, 2021.

\bibitem{ghamisi2018}
P.~Ghamisi and N.~Yokoya, ``{{IMG2DSM}}: {{Height Simulation From Single
  Imagery Using Conditional Generative Adversarial Net}},'' \emph{IEEE
  Geoscience and Remote Sensing Letters}, vol.~15, no.~5, pp. 794--798, May
  2018.

\bibitem{paoletti2021}
M.~E. Paoletti, J.~M. Haut, P.~Ghamisi, N.~Yokoya, J.~Plaza, and A.~Plaza,
  ``U-{{IMG2DSM}}: {{Unpaired Simulation}} of {{Digital Surface Models With
  Generative Adversarial Networks}},'' \emph{IEEE Geoscience and Remote Sensing
  Letters}, vol.~18, no.~7, pp. 1288--1292, Jul. 2021.

\bibitem{he2018}
K.~He, G.~Gkioxari, P.~Doll{\'a}r, and R.~Girshick, ``Mask {{R-CNN}},'' Jan.
  2018.

\bibitem{mahmud2020}
J.~Mahmud, T.~Price, A.~Bapat, and J.-M. Frahm, ``Boundary-{{Aware 3D Building
  Reconstruction From}} a {{Single Overhead Image}},'' in \emph{2020
  {{IEEE}}/{{CVF Conference}} on {{Computer Vision}} and {{Pattern
  Recognition}} ({{CVPR}})}.\hskip 1em plus 0.5em minus 0.4em\relax {Seattle,
  WA, USA}: {IEEE}, Jun. 2020, pp. 438--448.

\bibitem{chen2021}
S.~Chen, L.~Mou, Q.~Li, Y.~Sun, and X.~X. Zhu, ``Mask-{{Height R-CNN}}: {{An
  End-to-End Network}} for {{3D Building Reconstruction}} from {{Monocular
  Remote Sensing Imagery}},'' in \emph{2021 {{IEEE International Geoscience}}
  and {{Remote Sensing Symposium IGARSS}}}, Jul. 2021, pp. 1202--1205.

\bibitem{sun2022a}
W.~Sun, Y.~Zhang, Y.~Liao, B.~Yang, M.~Lin, R.~Zhai, and Z.~Gao, ``Rethinking
  {{Monocular Height Estimation}} from a {{Classification Task Perspective
  Leveraging}} the {{Vision Transformer}},'' \emph{IEEE Geoscience and Remote
  Sensing Letters}, pp. 1--1, 2022.

\bibitem{li2023}
Q.~Li, L.~Mou, Y.~Hua, Y.~Shi, S.~Chen, Y.~Sun, and X.~X. Zhu,
  ``{3DCentripetalNet}: {Building Height Retrieval} from {Monocular Remote
  Sensing Imagery},'' \emph{International Journal of Applied Earth Observation
  and Geoinformation}, 2023, in press.

\bibitem{eigen2014}
D.~Eigen, C.~Puhrsch, and R.~Fergus, ``Depth {{Map Prediction}} from a {{Single
  Image}} using a {{Multi-Scale Deep Network}},'' \emph{arXiv:1406.2283 [cs]},
  Jun. 2014.

\bibitem{eigen2015}
D.~Eigen and R.~Fergus, ``Predicting {{Depth}}, {{Surface Normals}} and
  {{Semantic Labels}} with a {{Common Multi-scale Convolutional
  Architecture}},'' in \emph{2015 {{IEEE International Conference}} on
  {{Computer Vision}} ({{ICCV}})}.\hskip 1em plus 0.5em minus 0.4em\relax
  {Santiago, Chile}: {IEEE}, Dec. 2015, pp. 2650--2658.

\bibitem{zhang2018}
Z.~Zhang, Z.~Cui, C.~Xu, Z.~Jie, X.~Li, and J.~Yang, ``Joint {{Task-Recursive
  Learning}} for {{Semantic Segmentation}} and {{Depth Estimation}},'' in
  \emph{Computer {{Vision}} \textendash{} {{ECCV}} 2018}, V.~Ferrari,
  M.~Hebert, C.~Sminchisescu, and Y.~Weiss, Eds.\hskip 1em plus 0.5em minus
  0.4em\relax {Cham}: {Springer International Publishing}, 2018, vol. 11214,
  pp. 238--255.

\bibitem{lore2018}
K.~G. Lore, K.~Reddy, M.~Giering, and E.~A. Bernal, ``Generative {{Adversarial
  Networks}} for {{Depth Map Estimation}} from {{RGB Video}},'' in \emph{2018
  {{IEEE}}/{{CVF Conference}} on {{Computer Vision}} and {{Pattern Recognition
  Workshops}} ({{CVPRW}})}.\hskip 1em plus 0.5em minus 0.4em\relax {Salt Lake
  City, UT}: {IEEE}, Jun. 2018, pp. 1258--12\,588.

\bibitem{fu2018}
H.~Fu, M.~Gong, C.~Wang, K.~Batmanghelich, and D.~Tao, ``Deep {{Ordinal
  Regression Network}} for {{Monocular Depth Estimation}},'' in \emph{2018
  {{IEEE}}/{{CVF Conference}} on {{Computer Vision}} and {{Pattern
  Recognition}}}.\hskip 1em plus 0.5em minus 0.4em\relax {Salt Lake City, UT}:
  {IEEE}, Jun. 2018, pp. 2002--2011.

\bibitem{li2020}
X.~Li, M.~Wang, and Y.~Fang, ``Height {{Estimation From Single Aerial Images
  Using}} a {{Deep Ordinal Regression Network}},'' \emph{IEEE Geoscience and
  Remote Sensing Letters}, pp. 1--5, 2020.

\bibitem{li2022}
Z.~Li, X.~Wang, X.~Liu, and J.~Jiang, ``Binsformer: Revisiting adaptive bins
  for monocular depth estimation,'' \emph{arXiv preprint arXiv:2204.00987},
  2022.

\bibitem{bhat2022}
S.~F. Bhat, I.~Alhashim, and P.~Wonka, ``{{LocalBins}}: {{Improving Depth
  Estimation}} by {{Learning Local Distributions}},'' Mar. 2022.

\bibitem{fan2016}
H.~Fan, H.~Su, and L.~Guibas, ``A {{Point Set Generation Network}} for {{3D
  Object Reconstruction}} from a {{Single Image}},'' Dec. 2016.

\bibitem{bosch2019}
M.~Bosch, K.~Foster, G.~Christie, S.~Wang, G.~D. Hager, and M.~Brown,
  ``Semantic {{Stereo}} for {{Incidental Satellite Images}},'' in \emph{2019
  {{IEEE Winter Conference}} on {{Applications}} of {{Computer Vision}}
  ({{WACV}})}.\hskip 1em plus 0.5em minus 0.4em\relax {Waikoloa Village, HI,
  USA}: {IEEE}, Jan. 2019, pp. 1524--1532.

\bibitem{christie2020}
G.~Christie, R.~R.~R. Munoz~Abujder, K.~Foster, S.~Hagstrom, G.~D. Hager, and
  M.~Z. Brown, ``Learning {{Geocentric Object Pose}} in {{Oblique Monocular
  Images}},'' in \emph{2020 {{IEEE}}/{{CVF Conference}} on {{Computer Vision}}
  and {{Pattern Recognition}} ({{CVPR}})}.\hskip 1em plus 0.5em minus
  0.4em\relax {Seattle, WA, USA}: {IEEE}, Jun. 2020, pp. 14\,500--14\,508.

\bibitem{christie2021}
G.~Christie, K.~Foster, S.~Hagstrom, G.~D. Hager, and M.~Z. Brown, ``Single
  {{View Geocentric Pose}} in the {{Wild}},'' in \emph{2021 {{IEEE}}/{{CVF
  Conference}} on {{Computer Vision}} and {{Pattern Recognition Workshops}}
  ({{CVPRW}})}.\hskip 1em plus 0.5em minus 0.4em\relax {Nashville, TN, USA}:
  {IEEE}, Jun. 2021, pp. 1162--1171.

\bibitem{lesaux2019}
B.~Le~Saux, N.~Yokoya, R.~Haensch, and M.~Brown, ``2019 {{IEEE GRSS Data Fusion
  Contest}}: {{Large-Scale Semantic 3D Reconstruction}} [{{Technical
  Committees}}],'' \emph{IEEE Geoscience and Remote Sensing Magazine}, vol.~7,
  no.~4, pp. 33--36, Dec. 2019.

\bibitem{isprs}
``{{2D Semantic Label}}. - {{Vaihingen}},''
  https://www.isprs.org/education/benchmarks/UrbanSemLab/2d-sem-label-vaihingen.aspx.

\bibitem{gerke2014}
M.~Gerke, ``Use of the stair vision library within the {{ISPRS 2D}} semantic
  labeling benchmark ({{Vaihingen}}),'' Dec. 2014.

\bibitem{tan2020}
M.~Tan and Q.~V. Le, ``{{EfficientNet}}: {{Rethinking Model Scaling}} for
  {{Convolutional Neural Networks}},'' Sep. 2020.

\bibitem{alhashim2019}
I.~Alhashim and P.~Wonka, ``High {{Quality Monocular Depth Estimation}} via
  {{Transfer Learning}},'' Mar. 2019.

\end{thebibliography}

\begin{IEEEbiography}[{\includegraphics[width=1in,height=1.25in,clip,keepaspectratio]{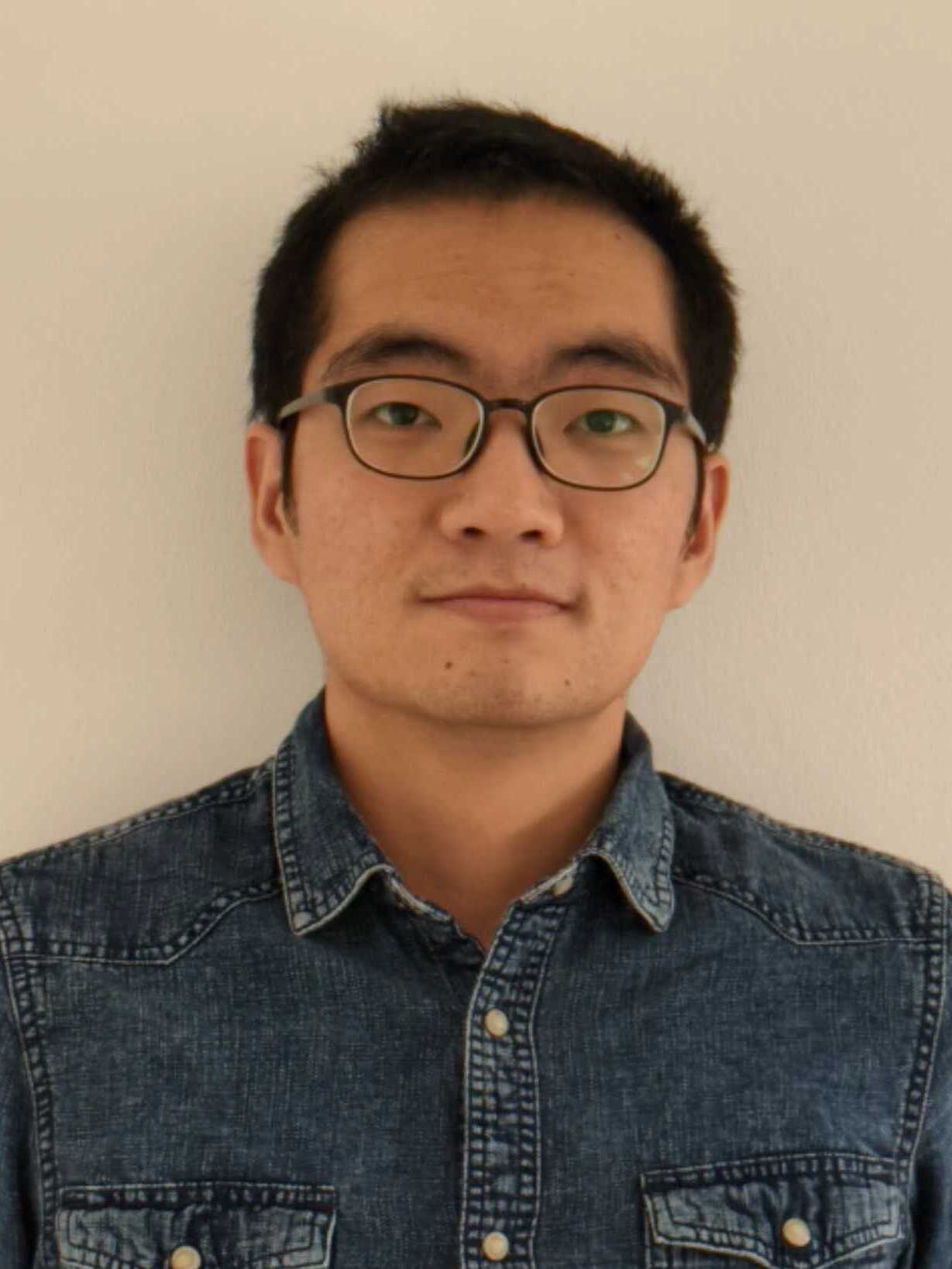}}]{Sining Chen}received the Bachelor's degree in marine science at Xiamen University, Xiamen, China, in 2018, and the Master's degree in Earth-oriented Space Science and Technology (ESPACE) at Technical University of Munich (TUM), Munich, Germany in 2020. He is pursuing a Ph.D. degree at the Chair of Data Science in Earth Observation at Technical University of Munich (TUM) since September 2021. He was a DLR/DAAD Doctoral Research Fellow at the Remote Sensing Technology Institute, German Aerospace Center (DLR), Wessling, Germany, from September 2021 to August 2023. His research interests include deep learning, monocular height estimation, and 3D building reconstruction.
\end{IEEEbiography}

\begin{IEEEbiography}[{\includegraphics[width=1in,height=1.25in,clip,keepaspectratio]{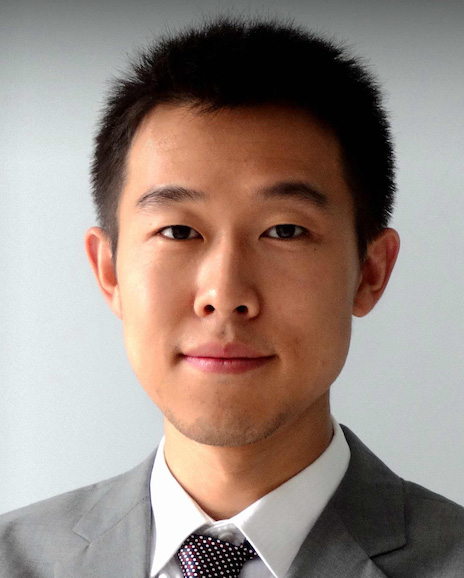}}]{Yilei Shi} (Member, IEEE) received the Dipl.-Ing. degree in mechanical engineering and the Dr.-Ing. degree in signal processing from the Technical University of Munich (TUM), Munich, Germany, in 2010 and 2019, respectively. He is a Senior Scientist with the Chair of Remote Sensing Technology, TUM. His research interests include fast solver and parallel computing for large-scale problems, high-performance computing and computational intelligence, advanced methods on synthetic aperture radar (SAR) and InSAR processing, machine learning, and deep learning for variety of data sources, such as SAR, optical images, and medical images, and partial differential equation (PDE)-related numerical modeling and computing.
\end{IEEEbiography}

\begin{IEEEbiography}[{\includegraphics[width=1in,height=1.25in,clip,keepaspectratio]{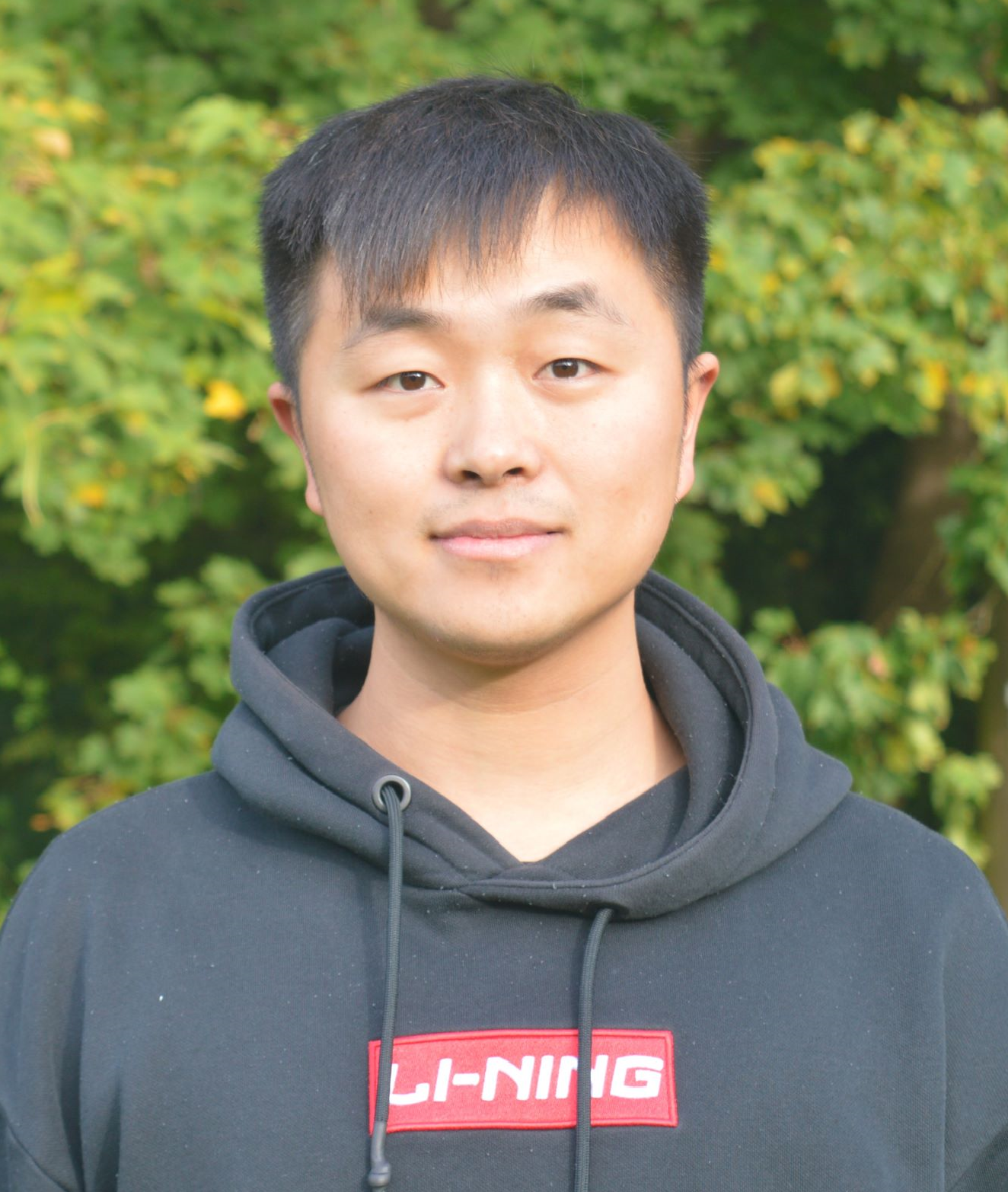}}]{Zhitong Xiong} (Member, IEEE) received the Ph.D. degree in computer science and technology from Northwestern Polytechnical University, Xi’an, China, in 2021. He is currently a research scientist and leads the ML4Earth working group with the Data Science in Earth Observation, Technical University of Munich (TUM), Germany. His research interests include computer vision, machine learning, Earth observation, and Earth system modeling.
\end{IEEEbiography}

\begin{IEEEbiography}[{\includegraphics[width=1in,height=1.25in,clip,keepaspectratio]{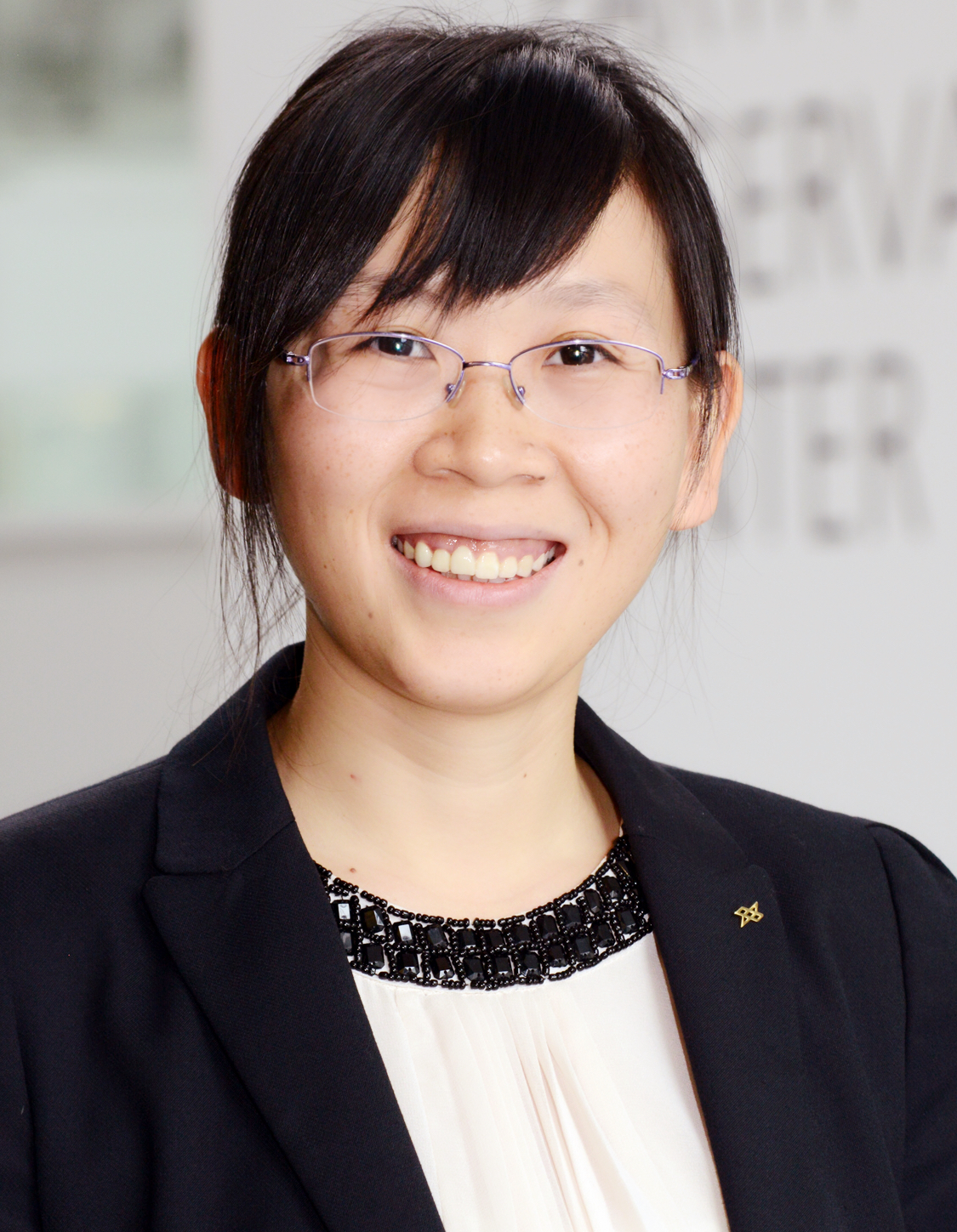}}]{Xiao Xiang Zhu}(S'10--M'12--SM'14--F'21) received the Master (M.Sc.) degree, her doctor of engineering (Dr.-Ing.) degree and her “Habilitation” in the field of signal processing from Technical University of Munich (TUM), Munich, Germany, in 2008, 2011 and 2013, respectively.
\par
She is the Chair Professor for Data Science in Earth Observation at Technical University of Munich (TUM) and was the founding Head of the Department ``EO Data Science'' at the Remote Sensing Technology Institute, German Aerospace Center (DLR). Since May 2020, she is the PI and director of the international future AI lab "AI4EO -- Artificial Intelligence for Earth Observation: Reasoning, Uncertainties, Ethics and Beyond", Munich, Germany. Since October 2020, she also serves as a Director of the Munich Data Science Institute (MDSI), TUM. From 2019 to 2022, Zhu has been a co-coordinator of the Munich Data Science Research School (www.mu-ds.de) and the head of the Helmholtz Artificial Intelligence -- Research Field ``Aeronautics, Space and Transport".  Prof. Zhu was a guest scientist or visiting professor at the Italian National Research Council (CNR-IREA), Naples, Italy, Fudan University, Shanghai, China, the University  of Tokyo, Tokyo, Japan and University of California, Los Angeles, United States in 2009, 2014, 2015 and 2016, respectively. She is currently a visiting AI professor at ESA's Phi-lab. Her main research interests are remote sensing and Earth observation, signal processing, machine learning and data science, with their applications in tackling societal grand challenges, e.g. Global Urbanization, UN’s SDGs and Climate Change.

Dr. Zhu has been a member of young academy (Junge Akademie/Junges Kolleg) at the Berlin-Brandenburg Academy of Sciences and Humanities and the German National  Academy of Sciences Leopoldina and the Bavarian Academy of Sciences and Humanities. She serves in the scientific advisory board in several research organizations, among others the German Research Center for Geosciences (GFZ, 2020-2023) and Potsdam Institute for Climate Impact Research (PIK). She is an associate Editor of IEEE Transactions on Geoscience and Remote Sensing, Pattern Recognition and serves as the area editor responsible for special issues of IEEE Signal Processing Magazine. She is a Fellow of IEEE.
\end{IEEEbiography}




\end{document}